\documentclass[10pt,twocolumn,letterpaper]{article}
\usepackage[pagenumbers]{cvpr}
\usepackage{iftex}
\ifPDFTeX\pdfminorversion=7\else\ifXeTeX\fi\fi
\usepackage[T1]{fontenc}
\usepackage{microtype}
\usepackage{multirow,float,placeins}
\definecolor{cvprblue}{rgb}{0.21,0.49,0.74}
\usepackage[pagebackref,breaklinks,colorlinks,hypertexnames=false,allcolors=cvprblue]{hyperref}
\hypersetup{pdftitle={SWIFT: Slow World, Instant Freshness, and Timely Action for Asynchronous Autonomous Driving},pdfauthor={Jiazhe Wang, Tianyi Yang, Ruihang Li, Fan Lu, Xianming Zeng, Jianyun Xu, Xingliang Liu, Lei Yang}}
\newcommand{\ours}{SWIFT\xspace}
\providecommand{\Na}{\ensuremath{N_{\mathrm{a}}}}
\providecommand{\Hb}{\ensuremath{H_{\mathrm{b}}}}

\providecommand{\rtime}{\ensuremath{\varrho}}
\providecommand{\fa}{\ensuremath{f_{\mathrm{a}}}}
\providecommand{\fv}{\ensuremath{f_{\mathrm{v}}}}
\providecommand{\taud}{\ensuremath{\tau_{\mathrm{d}}}}
\providecommand{\Tpred}{\ensuremath{T_{\mathrm{p}}}}
\providecommand{\Tupd}{\ensuremath{T_{\mathrm{u}}}}
\providecommand{\tauw}{\ensuremath{\tau_{\mathrm{w}}}}
\providecommand{\taua}{\ensuremath{\tau_{\mathrm{a}}}}

\providecommand{\Plan}{\ensuremath{\hat{P}_{t}}}

\providecommand{\CurImgs}{\ensuremath{\mathcal{I}_{t}}}

\providecommand{\Wtheta}{\ensuremath{\mathcal{W}_{\theta}}}
\providecommand{\EVAE}{\ensuremath{E_{\mathrm{VAE}}}}
\providecommand{\sigv}{\ensuremath{\sigma_{\mathrm{v}}}}
\providecommand{\epsv}{\ensuremath{\epsilon_{\mathrm{v}}}}
\providecommand{\Znoisy}{\ensuremath{Z_{\sigma_{\mathrm{v}}}}}
\providecommand{\Nw}{\ensuremath{N_{\mathrm{w}}}}
\providecommand{\Nh}{\ensuremath{N_{\mathrm{h}}}}
\providecommand{\dhead}{\ensuremath{d_{\mathrm{h}}}}

\providecommand{\Mw}{\ensuremath{M_{\mathrm{w}}}}
\providecommand{\vTheta}{\ensuremath{v_{\theta}}}
\providecommand{\Etxt}{\ensuremath{E_{\mathrm{txt}}}}

\providecommand{\Nf}{\ensuremath{N_{\mathrm{f}}}}
\providecommand{\Ecar}{\ensuremath{E_{\mathrm{car}}}}
\providecommand{\dcar}{\ensuremath{d_{\mathrm{car}}}}
\providecommand{\Epos}{\ensuremath{E_{\mathrm{pos}}}}
\providecommand{\Wfin}{\ensuremath{W_{\mathrm{f},\mathrm{in}}}}
\providecommand{\WfK}{\ensuremath{W_{\mathrm{f},K}}}
\providecommand{\WfV}{\ensuremath{W_{\mathrm{f},V}}}
\providecommand{\Zf}{\ensuremath{Z_{\mathrm{f},t}}}

\providecommand{\sigmax}{\ensuremath{\sigma_{\max}}}
\providecommand{\sigmin}{\ensuremath{\sigma_{\min}}}
\providecommand{\NODE}{\ensuremath{N_{\mathrm{ODE}}}}
\providecommand{\Qa}{\ensuremath{Q_{\mathrm{a}}}}
\providecommand{\Ka}{\ensuremath{K_{\mathrm{a}}}}
\providecommand{\Va}{\ensuremath{V_{\mathrm{a}}}}
\providecommand{\Oa}{\ensuremath{O_{\mathrm{a}}}}
\providecommand{\Ma}{\ensuremath{M_{\mathrm{a}}}}

\providecommand{\SEtwo}{\ensuremath{\mathrm{SE}(2)}}

\providecommand{\Nvoc}{\ensuremath{N_{\mathrm{voc}}}}
\providecommand{\Nnoise}{\ensuremath{N_{\mathrm{noise}}}}

\providecommand{\lossTot}{\ensuremath{\mathcal{L}}}
\providecommand{\lossV}{\ensuremath{\mathcal{L}_{\mathrm{v}}}}
\providecommand{\lossA}{\ensuremath{\mathcal{L}_{\mathrm{a}}}}
\providecommand{\lossTraj}{\ensuremath{\mathcal{L}_{\mathrm{traj}}}}
\providecommand{\lossS}{\ensuremath{\mathcal{L}_{\mathrm{sc}}}}

\providecommand{\Fw}{\ensuremath{F_{\mathrm{w}}}}
\providecommand{\Fa}{\ensuremath{F_{\mathrm{a}}}}
\providecommand{\Fcar}{\ensuremath{F_{\mathrm{car}}}}
\providecommand{\Fsc}{\ensuremath{F_{\mathrm{sc}}}}
\providecommand{\Fdec}{\ensuremath{F_{\mathrm{dec}}}}

\providecommand{\RoPE}{\ensuremath{\operatorname{RoPE}}}

\providecommand{\softmax}{\ensuremath{\operatorname{softmax}}}
\providecommand{\BCE}{\ensuremath{\operatorname{BCE}}}

\providecommand{\argmax}{\operatorname*{arg\,max}}

\providecommand{\FlowUniPC}{FlowUniPC\xspace}
\providecommand{\WanVAE}{Wan-VAE\xspace}
\providecommand{\TAEW}{TAEW2.1\xspace}
\providecommand{\DINOvthree}{DINOv3\xspace}
\providecommand{\TransFuserName}{TransFuser\xspace}

\title{SWIFT: Slow World, Instant Freshness, and Timely Action for Asynchronous Autonomous Driving}
\author{\begin{tabular}{c}
Jiazhe Wang$^{1}$\quad Tianyi Yang$^{1}$\quad Ruihang Li$^{2}$\quad Fan Lu$^{3}$\\[4pt]
Xianming Zeng$^{1}$\quad Jianyun Xu$^{1,\dagger,*}$\quad Xingliang Liu$^{1,*}$\quad Lei Yang$^{1,*}$\\[8pt]
\small $^{1}$Hello Robotaxi\quad $^{2}$Zhejiang University\quad $^{3}$Tongji University\\[4pt]
\small $^{\dagger}$Project leader\qquad $^{*}$Corresponding authors
\end{tabular}}
\date{}
\begin{document}
\maketitle
\begin{abstract}
Driving scenes keep changing during model computation, making both observation freshness and action-delivery time central to world-action planning.
To retain rich generative world priors in autonomous driving without stalling real-time control, we present \textbf{\ours{}} (\textbf{S}low \textbf{W}orld, \textbf{I}nstant \textbf{F}reshness, and \textbf{T}imely action), a deployment-oriented asynchronous world-action system that moves world updates off the observation-to-action critical path.
On a background GPU, a slow Video DiT streams surround-view history into read-only predictive world context, trained with a predictive span that covers both the update interval and background world computation.
On a foreground GPU, a lightweight freshness carrier encodes current surround-view images into a gated key/value bank at every step, letting a fast Action DiT combine live visual feedback with the latest completed world context without waiting.
On open-loop NAVSIM, \ours{} reaches 94.2 PDMS at 198.4\,ms per decision with a trajectory scorer; on closed-loop Bench2Drive, it achieves an 80.30 Driving Score at 109.9\,ms foreground latency.
\ours{} further transfers to the Waymo E2E hidden test (8.0050 Rater Feedback Score) and nuScenes (0.54\,m mean endpoint error, 0.03\% mean collision rate).
Across all four benchmarks, \ours{} achieves the best results among compared world-action methods, establishing a favorable accuracy--latency trade-off.
\end{abstract}

\begin{figure}[!htbp]
  \centering
  \includegraphics[width=\linewidth]{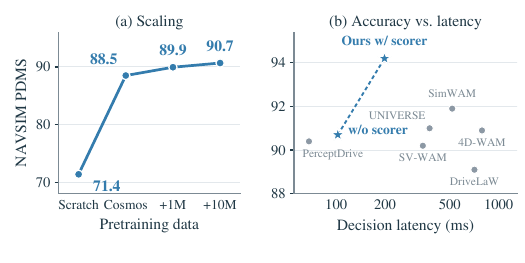}
  \input{figures/captions/fig_pretraining_transfer_caption.tex}
\end{figure}
\section{Introduction}
\label{sec:intro}

End-to-end autonomous driving requires both rich physical foresight to anticipate multi-agent interactions and fast closed-loop responsiveness to handle sudden scene changes across all surrounding views.
Conventional end-to-end planners typically learn scene representations from costly 3D bounding-box, HD-map, or occupancy annotations, making large-scale training difficult to sustain.
World-action models (WAMs)~\cite{driveva,lingbotva,fastwam,dreamzero} alleviate this annotation bottleneck by coupling generative world modeling~\cite{gaia1,vista,wan} with action prediction, learning physical world dynamics directly from multi-camera video and logged ego trajectories.
Scaling such generative pretraining substantially strengthens trajectory planning: on NAVSIM~\cite{navsim} \texttt{navtest}, our generator (without scoring) improves from 71.4 PDM Score (PDMS) when trained from scratch to 88.5 with public Cosmos-Predict2.5~\cite{cosmos_predict25} weights, 89.9 after joint world-action pretraining on 1M driving clips, and 90.7 on 10M clips disjoint from NAVSIM (\cref{fig:pretraining-transfer}a), matching scaling trends in driving world models~\cite{vavim,drivelaw}.

Deploying large-scale WAMs in real-time driving, however, exposes a fundamental timing conflict because traffic continues to evolve while the model computes (\cref{fig:world_action_inference}, top).
Most existing driving WAMs~\cite{drivewam,drivelaw,driveva,svwam,simwam} adopt a synchronous, chunk-based execution schedule that creates two bottlenecks.
First, placing a multi-billion-parameter video backbone on every decision's critical path locks action-delivery latency to world computation: in \cref{fig:pretraining-transfer}b, stronger driving WAMs incur 340--790\,ms per decision, whereas faster single-view variants sacrifice spatial coverage and accuracy.
Second, chunk-based video backbones refresh their world state only at chunk boundaries, so new camera frames arriving during chunk collection or world inference cannot influence intermediate decisions.
In surround-view driving across 6--7 camera streams, a single world update approaches $1$\,s; simply running world updates asynchronously~\cite{ahawam,zimablue,glancewam,dualwam} is therefore insufficient unless cached context remains predictive across its entire computation delay and reuse interval, while live visual changes from all views reach every control step within milliseconds without re-invoking the heavy video backbone.

To resolve both bottlenecks, we present \textbf{\ours{}} (\textbf{S}low \textbf{W}orld, \textbf{I}nstant \textbf{F}reshness, and \textbf{T}imely action), a deployment-oriented asynchronous world-action system for autonomous driving (\cref{fig:world_action_inference}, bottom; \cref{fig:predictive-context-overview}) that moves world updates off the observation-to-action critical path across two GPUs (\cref{fig:async-context-reuse}).
On the background GPU, a slow 2B Video DiT streams clean surround-view history into a read-only key/value (K/V) cache without test-time video decoding, trained with a predictive span that covers both the update interval and background world computation.
At every foreground step, a frozen lightweight freshness carrier encodes the latest multi-camera images into a gated K/V bank in under 20\,ms for instant freshness, which a fast 0.3B Action DiT reads alongside cached world context and ego conditions for timely action while the next world update is in flight.

\begin{figure}[!htbp]
  \centering
  \includegraphics[width=0.92\linewidth]{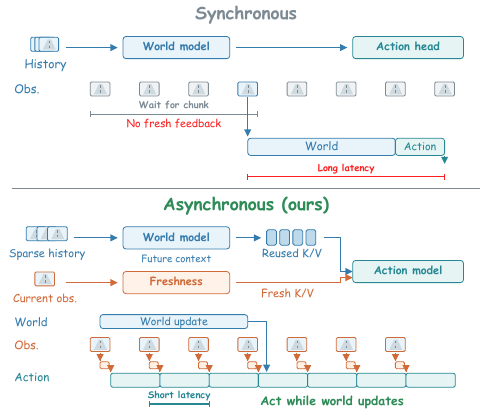}
  \input{figures/captions/fig_inference_paradigms_caption}
\end{figure}

Our main contributions are summarized as follows:
\begin{itemize}
    \item \textbf{\ours{}: an asynchronous world-action driving system.} We decouple streaming background world updates from fast foreground planning across two GPUs, moving surround-view world inference off the critical path while training the predictive span to cover the update interval plus measured update latency.
    \item \textbf{Lightweight freshness carrier for per-decision visual feedback.} A frozen multi-camera encoder with a gated K/V adapter injects live surround-view observations, effective with both compact reconstruction-based and semantic features, into every decision without modifying read-only world context or invoking the Video DiT.
    \item \textbf{Leading planning performance and delay-aware validation.} \ours{} leads compared world-action models on open-loop NAVSIM~\cite{navsim} (94.2 PDMS at 198.4\,ms with shared-feature candidate scoring), closed-loop Bench2Drive~\cite{bench2drive} (80.30 Driving Score at 109.9\,ms, validated under injected command delay), and the Waymo E2E~\cite{waymoe2e} hidden test (8.0050 Rater Feedback Score), running $4.7\times$ faster than synchronous execution at matched accuracy (\cref{fig:pretraining-transfer}b; \cref{tab:navsim,tab:b2d,tab:transfer-waymo,tab:latency-injection}).
\end{itemize}

\begin{figure*}[!t]
  \centering
  \includegraphics[width=0.94\textwidth]{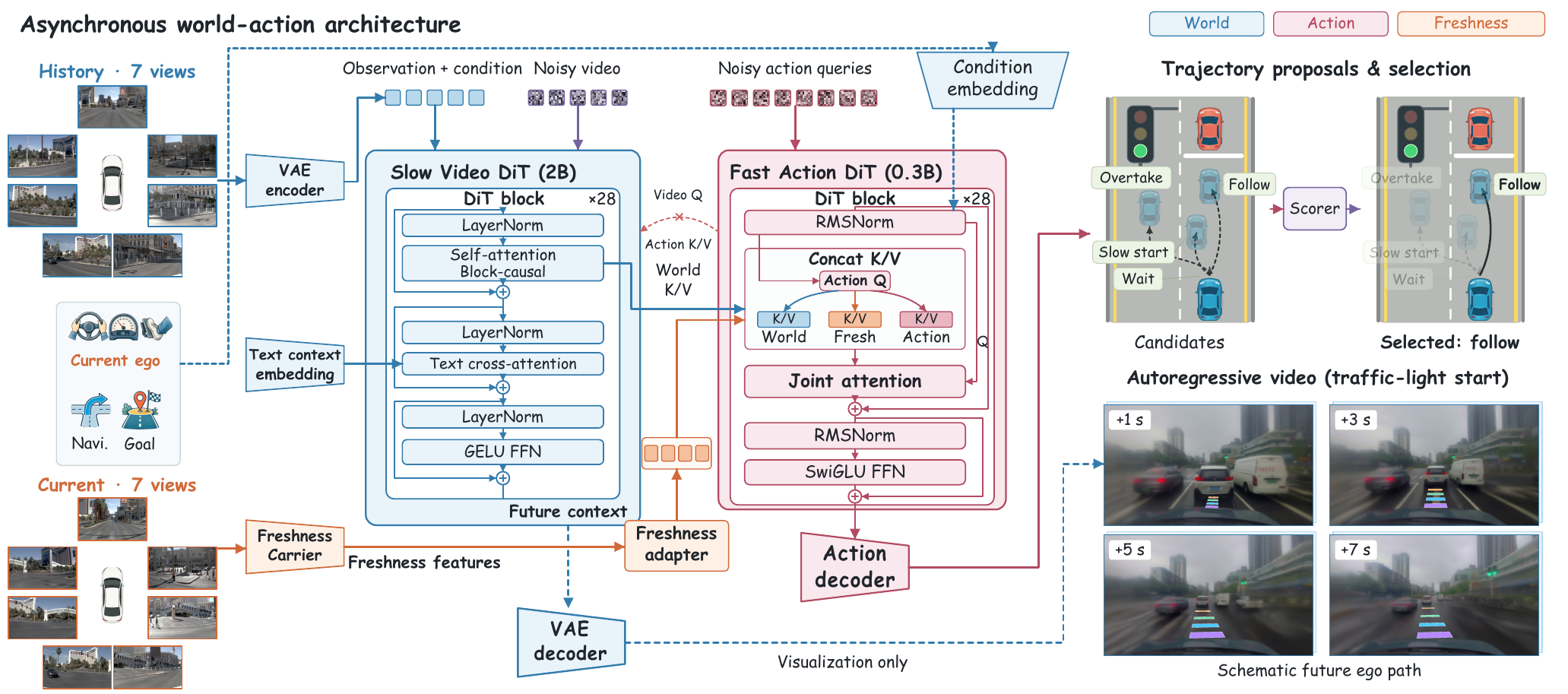}
  \input{figures/captions/fig_model_overview_caption}
\end{figure*}

\section{Related work}
\label{sec:related}

\subsection{World-action models}
World-action models (WAMs) couple visual dynamics with control to transfer video-pretraining priors into action prediction: LingBot-VA predicts future video latents before action decoding~\cite{lingbotva}, DreamZero jointly denoises video and actions~\cite{dreamzero}, and UniWAM jointly generates language, future video, and actions across coupled experts~\cite{uniwam}, whereas Fast-WAM uses video co-training solely to shape historical context without test-time video generation~\cite{fastwam}.
Although LingBot-VA and DreamZero pipeline chunk execution, all four still run a world pass per action chunk, tying reaction latency to world compute; we instead decouple background world updates from foreground planning while keeping visual freshness.

\subsection{End-to-end autonomous driving}
End-to-end driving planners map sensor inputs to trajectories via truncated diffusion~\cite{diffusiondrive}, proposal scoring~\cite{gtrs,drivesuprim,sparsedrivev2}, or 3D semantic alignment~\cite{rap}, yet typically rely on costly perception or HD-map labels.
Driving WAMs alleviate label requirements during representation learning via video priors: DriveWAM conditions planning on generated video~\cite{drivewam}, DriveVA and 4D-WAM jointly denoise video and trajectories~\cite{driveva,fourdwam}, DriveLaW reads intermediate video-denoising features~\cite{drivelaw}, and DriveVLA-W0, UNIVERSE, Metis, SV-WAM, and SimWAM bypass test-time video generation~\cite{drivevlaw0,universe,metis,svwam,simwam}.
Still, existing driving WAMs run their world backbone synchronously per decision; \ours{} scales perception-label-free pretraining (\cref{fig:pretraining-transfer}a) under asynchronous world updates, and its scorer reuses cached world and freshness features without a separate visual encoder.

\subsection{Asynchronous slow--fast world-action systems}
Dual-system policies decouple slow reasoning from fast control~\cite{gr00tn1,robodual}, while asynchronous execution, streaming perception, and action chunking account for inference delay~\cite{hirt,ticvla,streamingperception,rtc}; in driving, DriveVLM-Dual passes low-rate trajectory guidance and AsyncDriver supplies language guidance, rather than dense spatiotemporal priors~\cite{drivevlm,asyncdriver}.
Among manipulation WAMs over 1--3 views, AHA-WAM adapts cached K/V via observation queries~\cite{ahawam}, GlanceWAM pairs a held lookahead latent with per-step Video DiT encoding~\cite{glancewam}, DualWAM refines a partial plan with a wrist WAM~\cite{dualwam}, and ZimaBlue lets a fast DiT read K/V cached from a video-action rollout~\cite{zimablue}.
For surround-view driving, \ours{} differs in three ways (\cref{sec:supp-async-comparison}): unlike causal video diffusion~\cite{causvid,selfforcing}, its slow branch streams clean history into a read-only K/V cache without test-time video generation or cache editing; current views enter via a frozen lightweight carrier with a gated K/V bank; and the predictive span covers the update interval plus measured update latency.

\section{Methodology}
\label{sec:supp-method}
\label{sec:method}

\ours{} (\cref{fig:predictive-context-overview}) combines generative world representations with current surround-view evidence: a \emph{slow world} Video DiT builds reusable predictive context (\cref{sec:method-world}), an \emph{instant freshness} carrier encodes the latest images (\cref{sec:method-action}), and a fast Action DiT attends to both for \emph{timely action} generation, with optional candidate scoring (\cref{sec:method-scorer}).

\begin{figure*}[!t]
  \centering
  \includegraphics[width=0.94\textwidth]{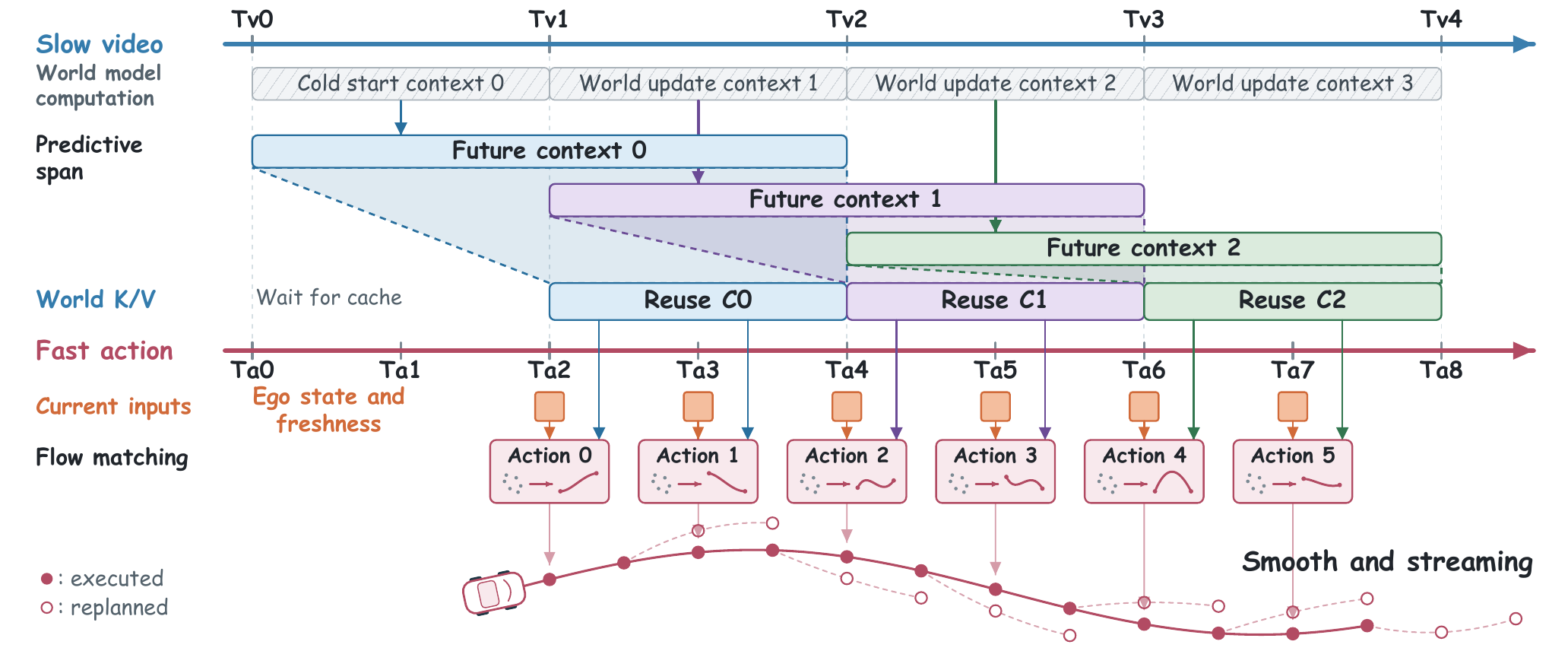}
  \input{figures/captions/fig_async_context_caption.tex}
\end{figure*}

\subsection{Predictive context and asynchronous reuse}
\label{sec:method-world}

\ours{} operates at two coupled timescales on two dedicated GPUs (\cref{fig:predictive-context-overview,fig:async-context-reuse}): the fast foreground (GPU 0) runs the Action DiT at frequency $\fa$ (decision interval $\Delta = 1/\fa$), and the slow background (GPU 1) runs the Video DiT, which encodes history as latent slots at a lower rate $\fv < \fa$ (slot spacing $1/\fv$).
At each decision time $t$, the foreground receives current surround-view images $\CurImgs=\{I_t^c\}_{c=1}^{C}$ across $C$ cameras and an ego-condition vector $e_t\in\mathbb R^{d_e}$ (encoding current kinematics and navigation goals), and outputs an ego-frame trajectory $\Plan=[(x_j,y_j,\psi_j)]_{j=1}^{\Na}$ over $\Na$ future waypoints spaced by $\Delta$.
Meanwhile, the background maintains $H$ observed latent slots at rate $\fv$ up to snapshot $s \le t$. Each world update adds one latent slot, so the world update interval is $\Tupd=1/\fv$. Here $\Hb$ separately denotes the number of future target slots used for predictive training.
Let $Z_s^{\mathrm{obs}}$ be the newest latent slot produced by the frozen causal VAE $\EVAE$~\cite{wan} from incoming camera frames using its streaming state, and let $\mathcal S_{s-\Tupd}$ denote the preceding world state. An $L=28$-layer, $2\text{B}$ Video DiT $\Wtheta$~\cite{cosmos_predict25} ($\Nh=16$ heads of dimension $\dhead=128$) updates this state using a mask $\Mw$ that lets each clean observation slot attend to itself and earlier clean slots. Spatial tokens within a slot attend bidirectionally.
\begin{equation}
(\mathcal C_s,X_s)=\Wtheta\big(Z_s^{\mathrm{obs}};\mathcal S_{s-\Tupd},\Etxt,\sigv\equiv0\big),
\label{eq:m-world}
\end{equation}
where $\Etxt$ is a fixed empty-prompt embedding, $\mathcal C_s=\{(K_{\mathrm w,s}^{\ell},V_{\mathrm w,s}^{\ell})\}_{\ell=1}^{L}$ is the cached per-layer world K/V bank, and $X_s\in\mathbb R^{\Nw\times D}$ ($D=2048$) are the retained final-layer residual-stream token features (world residuals). Here $\Nw$ counts tokens in the $H$ retained observation slots, and $\mathcal S_s=(\mathcal C_s,X_s)$ is the published world state.
During training, clean observation slots (video noise level $\sigv=0$) use the same single-slot causal history visibility. A future target block contains $\Hb$ noised slots ($\sigv>0$), which may attend to clean history and to one another, while clean history cannot attend to future targets. A video condition mask flags clean history as in Cosmos-Predict2.5~\cite{cosmos_predict25}. The target block sets the predictive span $\Tpred = \Hb/\fv$ independently of how many frames each world update adds.
At inference, $\Wtheta$ processes only clean observations ($\sigv\equiv0$) without video decoding, so $(\mathcal C_s,X_s)$ carries features trained to anticipate the next $\Tpred$ without generating it~\cite{fastwam} (\cref{sec:supp-world}).

\subsubsection{World attention and streaming context}
\label{sec:supp-world}
Per-head world self-attention at layer $\ell$ follows
\begin{equation}
O_{\mathrm w,s}^{\ell}=\softmax\!\left(\frac{Q_{\mathrm w,s}^{\ell}(K_{\mathrm w,s}^{\ell})^{\top}}{\sqrt{\dhead}}+\Mw\right)V_{\mathrm w,s}^{\ell},
\end{equation}
where $Q_{\mathrm w,s}^{\ell}\in\mathbb R^{N_q\times\dhead}$ and $K_{\mathrm w,s}^{\ell},V_{\mathrm w,s}^{\ell}\in\mathbb R^{N_k\times\dhead}$ with query and key token counts $N_q$ and $N_k$. During streaming updates, queries are formed solely from the newest latent slot whereas keys and values span that slot together with retained history, yielding $\mathcal C_s$.
During training, the model encodes finite $H$-slot clips under single-slot causal visibility with action gradients attached to $\mathcal C_s$.

\subsection{Asynchronous scheduling}
\label{sec:method-formulation}

Background world updates (every $\Tupd$, an integer multiple of $\Delta$) run concurrently with foreground decisions (every $\Delta$; \cref{fig:async-context-reuse}), so current observations reach planning throughout world computation.
After cold start, during which the background fills the $H$-slot cache and the ego vehicle stays stationary ($T_{\mathrm v0}\to T_{\mathrm v1}$ in the schematic \cref{fig:async-context-reuse}), the background launches a world update every $\Tupd$; its \emph{world update latency} $\tauw$ includes both the world model's computation and the communication between the two GPUs.
The background evicts the oldest slot and appends each new slot without modifying retained K/V. Cache updates persist across Bench2Drive decisions; training and NAVSIM evaluation instead encode independent history clips.
Once the update for a snapshot at $s$ completes at $s + \tauw$, its published context $(\mathcal C_s, X_s)$ serves $\Tupd/\Delta$ foreground decisions until the next update completes at $s + \Tupd + \tauw$.
Because each update is used as soon as it is ready and snapshots lie on the decision grid, the \emph{first-read delay} from a snapshot to its first use is fixed at $\taud = \lceil \tauw/\Delta \rceil\,\Delta$.
At decision time $t$, the Action DiT reads the latest completed cache without waiting. Its \emph{context age} follows
\begin{equation}
a_t=t-s\in\{\taud,\taud+\Delta,\ldots,\taud+\Tupd-\Delta\}.
\label{eq:context-age}
\end{equation}
\begin{table}[!htbp]
  \caption{\textbf{Timing of the asynchronous schedule.} The chunk step first reads a new world context ($a_t=\taud$); freshness steps follow.}
  \label{tab:timing}
  \centering
  \footnotesize
  \setlength{\tabcolsep}{3pt}
  \begin{tabular*}{\columnwidth}{@{\extracolsep{\fill}}llrr@{}}
    \toprule
    Quantity & Symbol & NAVSIM & Bench2Drive \\
    \midrule
    Decision interval & $\Delta$ & 0.5\,s & 0.1\,s \\
    World update interval & $\Tupd$ & 1.0\,s & 1.0\,s \\
    World update latency & $\tauw$ & 760.6\,ms & 941.5\,ms \\
    First-read delay & $\taud$ & 1.0\,s & 1.0\,s \\
    Context age & $a_t$ & 1.0, 1.5\,s & 1.0--1.9\,s \\
    Predictive span & $\Tpred$ & 2.0\,s & 2.0\,s \\
    Decision latency, w/o scorer & $\taua$ & 102.3\,ms & 109.9\,ms \\
    Decision latency, w/ scorer & $\taua$ & 198.4\,ms & -- \\
    \bottomrule
  \end{tabular*}
\end{table}

\Cref{tab:timing} lists all timing quantities on both benchmarks.
The decision that first reads a new cache, its \emph{chunk step}, sees $a_t = \taud$, and the following $\Tupd/\Delta - 1$ \emph{freshness steps} reuse the same cache until the next update arrives, receiving new observations only through the freshness carrier (\cref{sec:method-action}).

Since $\taud<\tauw+\Delta$, the last read of $(\mathcal C_s, X_s)$, at $s+\taud+\Tupd-\Delta$, precedes $s+\Tupd+\tauw$. Every read thus stays within the trained span, and world updates do not queue, if the first two conditions below hold; the third targets the foreground \emph{decision latency} $\taua$:
\begin{equation}
\Tpred \ge \Tupd + \tauw, \qquad \tauw \le \Tupd, \qquad \taua \le \Delta.
\label{eq:m-age}
\end{equation}
Bench2Drive exceeds the third by about 10\,ms, a control delay that \cref{tab:latency-injection} measures in closed loop.
With one slot per update ($\Tupd=1/\fv$) and $\Tpred=\Hb/\fv$, the first two reduce to $\Hb\ge1+\fv\tauw$ with $\fv\tauw\le1$, so every admissible latency yields the shortest span $\Hb=2$, which we use (\cref{tab:predictive-span}). This span covers scene evolution up to $t$, while the trajectory rollout over $[t, t+\Na\Delta]$ is grounded at $t$ by the freshness carrier and executed via receding-horizon control ($\bullet/\circ$ in \cref{fig:async-context-reuse}).

\begin{figure}[!tbp]
  \centering
  \includegraphics[width=\linewidth]{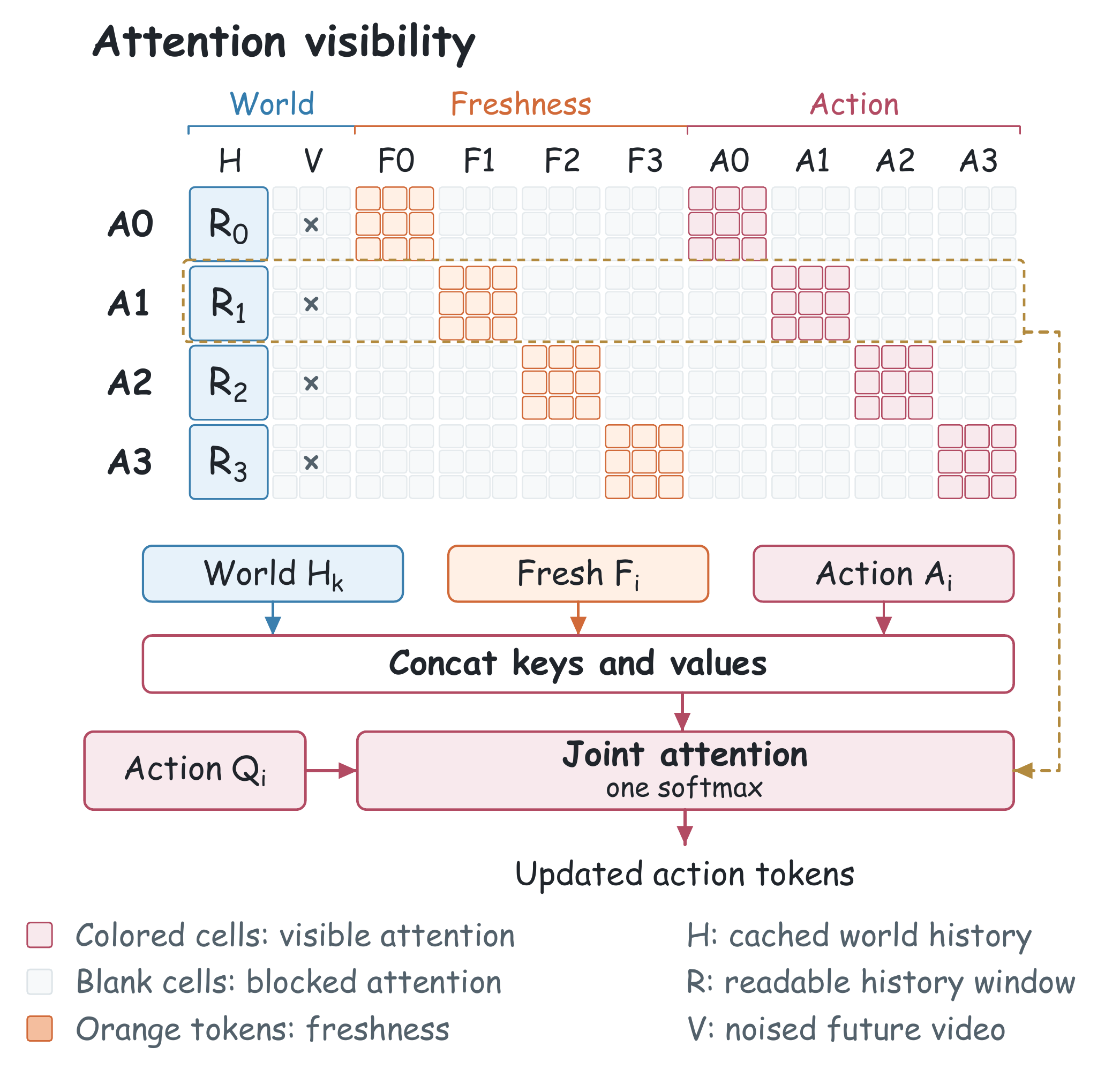}
  \input{figures/captions/fig_attention_visibility_caption.tex}
\end{figure}

\begin{figure*}[!t]
  \centering
  \includegraphics[width=0.88\textwidth]{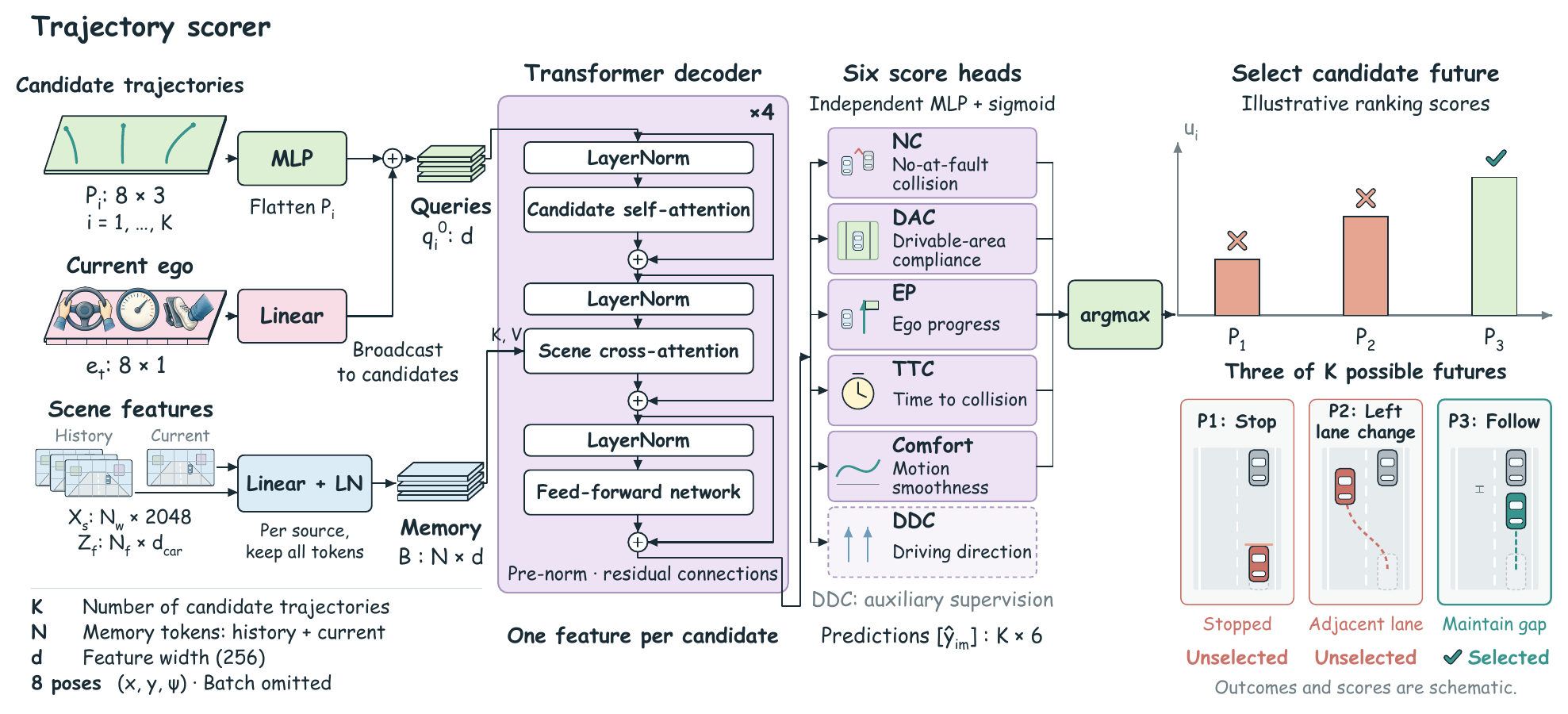}
  \input{figures/captions/fig3_caption.tex}
\end{figure*}

\subsection{Freshness-conditioned action generation}
\label{sec:method-action}

The world bank $\mathcal C_s$ summarizes observations up to $s$, while the freshness path admits current visual evidence independently of chunk collection and background world computation.
A frozen lightweight carrier $\Ecar$ (\eg, the tiny Wan2.1 autoencoder \TAEW~\cite{taehv}, \DINOvthree~\cite{dinov3}, or \TransFuserName~\cite{transfuser}) encodes all current views $\CurImgs$ at every step $t$ in 16.97--19.23\,ms without invoking the Video DiT.
A trainable adapter maps its patch features $\Zf=\Ecar(\CurImgs)\in\mathbb R^{\Nf\times\dcar}$ to a freshness K/V bank $\mathcal F_{t|s}=(K_{\mathrm f,t|s},\{V_{\mathrm f,t|s}^{\ell}\}_{\ell=1}^{L})$, whose keys are shared across layers and whose values are scaled per layer by zero-initialized gates $g^\ell$.
To encode context age $a_t = t - s$ without re-rotating cached world keys, retained world keys keep the absolute rotary positions assigned when they were encoded, while freshness and action tokens are rotated at continuous temporal coordinates from $\fv a_t$ slots after the newest cached slot~\cite{rope} (\cref{sec:supp-freshness-adapter}).

\subsubsection{Freshness adapter and temporal coordinates}
\label{sec:supp-freshness-adapter}
In the foreground, the freshness adapter projects $\Zf$ via $\Wfin\in\mathbb R^{\dcar\times512}$, adds camera-spatial embeddings $\Epos\in\mathbb R^{\Nf\times512}$, and applies layer-shared projections $\WfK,\WfV\in\mathbb R^{512\times D}$. Freshness keys are computed as $K_{\mathrm f,t|s}=\RoPE_{\rtime(t)}((\Zf\Wfin+\Epos)\WfK)$, while per-layer values $V_{\mathrm f,t|s}^{\ell}=g^{\ell}((\Zf\Wfin+\Epos)\WfV)$ are scaled by zero-initialized scalar gates $g^\ell$.
Retained world keys preserve their original absolute 3D RoPE coordinates after oldest-slot eviction: denoting the latest published slot index by $p_s$, freshness and action tokens are rotated at continuous temporal positions $\rtime(t)=p_s+\fv a_t$ and $\rtime(t+j\Delta)=p_s+\fv(a_t+j\Delta)$~\cite{rope}, so $\fv\taud=1$ places the chunk-step read exactly one slot after $p_s$.

\subsubsection{Action generation}
Within each of the $L=28$ blocks of the $0.3\text{B}$ Action DiT, noisy action tokens are modulated via AdaRMSNorm by the flow level $\sigma$ and $e_t$ and projected into $\Nh=16$ heads of width $\dhead=128$ (matching $\Wtheta$'s head geometry) to form $(\Qa^{\ell},\Ka^{\ell},\Va^{\ell})$, which perform joint attention over the concatenated world, freshness, and action tokens~\cite{pi0,fastwam}:
\begin{equation}
\begin{aligned}
\Oa^{\ell}={}&\softmax\!\left(
\frac{\Qa^{\ell}[K_{\mathrm w,s}^{\ell};K_{\mathrm f,t|s};\Ka^{\ell}]^{\top}}{\sqrt{\dhead}}+\Ma\right)\\
&\cdot[V_{\mathrm w,s}^{\ell};V_{\mathrm f,t|s}^{\ell};\Va^{\ell}],
\end{aligned}
\label{eq:m-attn}
\end{equation}
where $\Ma$ allows each action query to attend to $\mathcal C_s$, $\mathcal F_{t|s}$, and its own candidate group while masking other groups and, during training, all noised future video tokens~\cite{fastwam,svwam}.
World queries never attend to freshness or action K/V, keeping $\mathcal C_s$ read-only at inference.
A linear head predicts velocity $v_\eta\in\mathbb R^{\Na\times3}$ ($\eta$: Action DiT parameters) to obtain denoised waypoint increments $\hat A_0$, which a deterministic $\SEtwo$ rollout $P=G(\hat A_0)$ with exact inverse $G^{-1}$ accumulates into a trajectory (\cref{sec:supp-action}).

\subsubsection{Attention visibility and action decoding}
\label{sec:supp-action}

\Cref{fig:attention-visibility} illustrates the joint attention mask $\Ma$ in \cref{eq:m-attn} during training, where a single packed sequence spans multiple decisions (rows A0--A3). Specifically, under one unified softmax, each decision's action queries attend to the cached world-history window aligned with its own snapshot (R), its own freshness tokens (F), and its own action tokens (A), while masking out freshness and action tokens from all other decisions. Within each decision, the $K$ trajectory candidates share the same world and freshness tokens while forming mutually isolated action-token groups, each with internal bidirectional attention. Neither action queries nor clean-history queries attend to noised future-video tokens (V), and world queries never attend to freshness or action tokens, keeping the world K/V bank strictly read-only during action inference. At test time, only a single decision is evaluated per step, reducing $\Ma$ to the candidate-group mask in \cref{eq:m-attn}.

For trajectory decoding, the deterministic $\SEtwo$ rollout $G$ unscales predicted increments via $(\delta x_j,\delta y_j,\delta\psi_j)=\gamma\odot\hat A_{0,j}$ and integrates $(x_j,y_j)^{\top}=(x_{j-1},y_{j-1})^{\top}+R(\psi_{j-1})(\delta x_j,\delta y_j)^{\top}$ and $\psi_j=\psi_{j-1}+\delta\psi_j$ from $(0,0,0)$, where $R$ denotes planar rotation.

\subsection{Trajectory sampling and selection}
\label{sec:method-scorer}

To propose diverse candidates without re-encoding visual context, the Action DiT generates $K=\Nvoc+\Nnoise=53$ candidates $P_i$ in parallel in a single batched pass over shared $(\mathcal C_s,\mathcal F_{t|s})$.
The pool combines $\Nvoc=21$ trajectory priors with $\Nnoise=32$ Gaussian draws, denoised with one Euler step ($\NODE=1$); results without the scorer use one Gaussian draw (\cref{sec:supp-sampling}).

\subsubsection{Trajectory sampling}
\label{sec:supp-sampling}
Given noisy increments $A_\sigma\in\mathbb R^{\Na\times3}$, the Action DiT predicts $v_\eta(A_\sigma,\sigma;\mathcal C_s,\mathcal F_{t|s},e_t)$, and \FlowUniPC{}~\cite{flowmatching,unipc} at $\NODE=1$ simplifies to the single-step Euler update $\hat A_0=A_{\sigma_0}-\sigma_0v_\eta$.
Following anchored diffusion~\cite{diffusiondrive,gtrs}, multi-candidate inference noises $\Nvoc=21$ trajectory priors to $\sigma_0=0.5$ and combines them with $\Nnoise=32$ Gaussian draws at $\sigmax=0.999$, decoding each candidate via $P_i=G(\hat A_{0,i})$. The trajectory priors are used exclusively at inference, whereas training uses pure Gaussian noise.
Single-candidate ($K=1$) inference uses one Gaussian draw $\epsilon\sim\mathcal N(0,I)$ at $\sigma_0=0.999$ with one Euler step and no scorer.

\subsubsection{Trajectory scorer}
\label{sec:supp-scorer}
To select among candidates without a separate visual encoder, a lightweight Transformer scorer ($5.2\text{M}$ parameters)~\cite{gtrs,drivor,drivesuprim} embeds each $P_i$ with $e_t$ into a query and applies four decoder layers of candidate self-attention and cross-attention to a scene memory $B$ that concatenates detached world residuals $X_s$ with current freshness features $\Zf$.
Six sigmoid heads predict no at-fault collision (NC), drivable area compliance (DAC), ego progress (EP), time to collision (TTC), Comfort, and auxiliary driving direction compliance sub-scores $\hat y_{i,m}\in(0,1)$, and we select $\Plan=P_{\argmax_{i} u_i}$ with
\begin{equation}
u_i=\hat y_{i,\mathrm{NC}}\,\hat y_{i,\mathrm{DAC}}\,\tfrac{5\hat y_{i,\mathrm{TTC}}+5\hat y_{i,\mathrm{EP}}+2\hat y_{i,\mathrm{Comfort}}}{12},
\label{eq:m-select}
\end{equation}
which applies the PDMS aggregation to the predicted sub-score probabilities as a ranking surrogate for PDMS.
Fresh observations thus enter both candidate generation (via $\mathcal F_{t|s}$) and candidate scoring (via $\Zf$ in $B$; \cref{sec:supp-scorer}).

In \cref{fig:trajectory-scorer}, each candidate trajectory is embedded by an MLP and summed with a linear projection of the ego condition to form $d=256$ candidate queries $Q_{\mathrm{sc}}^0\in\mathbb R^{K\times d}$.
To form the dual-source scene memory, the scorer independently projects and LayerNorm-normalizes the $\Nw$ world residual tokens (width 2048) and $\Nf$ freshness tokens to width $d$, concatenating them into $B\in\mathbb R^{(\Nw+\Nf)\times d}$.
Four pre-norm Transformer decoder layers apply residual self-attention, cross-attention, and feed-forward blocks, followed by six heads: five sub-score heads aggregated via \cref{eq:m-select} and an auxiliary sixth head predicting driving-direction compliance.

\subsection{Joint training}
\label{sec:method-training}

During stage-one world-action training, $\Wtheta$, $v_\eta$, and the freshness adapter (together, the \emph{generator}) are optimized jointly while keeping $\EVAE$ and $\Ecar$ frozen.
With ground-truth increments $A_0=G^{-1}(P^\star)$ of the logged trajectory $P^\star$ and interpolated actions $A_\sigma=(1-\sigma)A_0+\sigma\epsilon$ for $\epsilon\sim\mathcal N(0,I)$ and $\sigma\in[\sigmin,1]$, the joint objective is:
\begin{equation}
\begin{aligned}
\lossA&=\mathbb E\big\|v_\eta(A_\sigma,\sigma;\mathcal C_s,\mathcal F_{t|s},e_t)
                    -(\epsilon-A_0)\big\|_2^2,\\
\lossTot&=\lambda_{\mathrm v}\lossV+\lambda_{\mathrm a}\lossA
               +\lambda_{\mathrm{traj}}\lossTraj ,
\end{aligned}
\label{eq:m-loss}
\end{equation}
where $\lossV$ is the rectified-flow loss on the noised future target slots~\cite{rectifiedflow,cosmos_predict25}, $\lossA$ is the action flow loss, and $\lossTraj$ is an $L_1$ waypoint loss on the decoded trajectory, with $(\lambda_{\mathrm v},\lambda_{\mathrm a},\lambda_{\mathrm{traj}})=(1,10,3)$.
During training, $\mathcal C_s$ remains attached to the gradient graph so action supervision directly refines $\Wtheta$ without feeding future trajectories as world inputs.

\subsubsection{Joint training objectives}
\label{sec:supp-training-objective}
In the joint training objective (\cref{eq:m-loss}), $\lossV=\mathbb E\|\vTheta(\Znoisy)-(\epsv-Z_0)\|_2^2$ supervises the Video DiT's velocity prediction $\vTheta$ on interpolated future target latents $\Znoisy=(1-\sigv)Z_0+\sigv\epsv$ with clean target latents $Z_0$, while $\lossTraj$ sums the mean and terminal waypoint $L_1$ errors of $G(A_\sigma-\sigma v_\eta)$.

In stage two, with the generator and carrier frozen, the scorer is trained over the generator's $K$-candidate pools on \texttt{navtrain} scenes and dual-source scene memory $B$ (from $X_s$ and $\Zf$) using $\lossS=\frac{1}{6K}\sum_{i,m}\BCE(\hat y_{i,m},y_{i,m})$ against sub-score targets $y_{i,m}$ computed by the NAVSIM simulator from annotated agent boxes and maps (\cref{sec:supp-training-objective,sec:supp-training-settings}).

\FloatBarrier
\section{Experiments}
\label{sec:exp}

\subsection{Setup}
\label{sec:exp-setup}
\label{sec:supp-impl}

\begin{table*}[!t]
  \caption{\textbf{NAVSIM v1 comparison} on \texttt{navtest}. In.: camera views (V) and LiDAR (L). Ann.: perception or HD-map label supervision, including derived simulator sub-scores. Latencies follow each paper's hardware ($\dagger$: our H20 rerun); ours is the foreground decision on one H20 while the world model updates on a second. \ours{} uses a 7\,s history and the chunk-step age (\cref{sec:exp-setup}). Bold: best accuracy per column; dashes: unavailable.}
  \label{tab:navsim}
  \centering
  \footnotesize
  \begin{tabular*}{\textwidth}{@{\extracolsep{\fill}}lcccccccccc@{}}
    \toprule
    Method & Year & In. & Ann. & NC $\uparrow$ & DAC $\uparrow$ & EP $\uparrow$ & TTC $\uparrow$ & Comfort $\uparrow$ & PDMS $\uparrow$ & \shortstack{Latency\\(ms) $\downarrow$} \\
    \midrule
    \multicolumn{11}{@{}l}{\emph{World-action methods}} \\
    DriveLaW~\cite{drivelaw} & 2026 & 1V & $\times$ & 99.0 & 97.1 & 81.3 & 96.7 & \textbf{100.0} & 89.1 & 710.0 \\
    DriveVLA-W0~\cite{drivevlaw0} & 2026 & 1V & $\times$ & 98.7 & 99.1 & 83.3 & 95.3 & 99.3 & 90.2 & -- \\
    SV-WAM~\cite{svwam} & 2026 & 6V & \checkmark & 98.6 & 98.6 & 83.9 & 95.1 & \textbf{100.0} & 90.2 & 341.6 \\
    PerceptDrive~\cite{perceptdrive} & 2026 & 1V & \checkmark & 99.2 & 99.0 & 82.9 & 98.0 & \textbf{100.0} & 90.4 & 68.0 \\
    4D-WAM~\cite{fourdwam} & 2026 & 1V & $\times$ & 99.1 & 97.9 & 84.6 & 96.6 & \textbf{100.0} & 90.9 & 790.0 \\
    DriveVA~\cite{driveva} & 2026 & 1V & $\times$ & 99.2 & 97.5 & 83.5 & \textbf{98.7} & \textbf{100.0} & 90.9 & -- \\
    UNIVERSE~\cite{universe} & 2026 & 1V & $\times$ & 99.1 & 97.6 & 83.6 & 98.5 & \textbf{100.0} & 91.0 & 376.0 \\
    SimWAM~\cite{simwam} & 2026 & 1V & \checkmark & 98.5 & 99.1 & 86.5 & 96.1 & 99.8 & 91.9 & 518.0 \\
    \midrule
    \multicolumn{11}{@{}l}{\emph{Other planners}} \\
    GoalFlow~\cite{goalflow} & 2025 & 3V+L & \checkmark & 98.4 & 98.3 & 85.0 & 94.6 & \textbf{100.0} & 90.3 & 532.1$^{\dagger}$ \\
    ReCogDrive~\cite{recogdrive} & 2026 & 1V & \checkmark & 97.9 & 97.3 & 87.3 & 94.9 & \textbf{100.0} & 90.8 & 75.0 \\
    Hydra-MDP++~\cite{hydramdppp} & 2025 & 3V & \checkmark & 98.6 & 98.6 & 85.7 & 95.1 & \textbf{100.0} & 91.0 & 271.0 \\
    DiffusionDriveV2~\cite{diffusiondrivev2} & 2025 & 3V+L & \checkmark & 98.3 & 97.9 & 87.5 & 94.8 & 99.9 & 91.2 & -- \\
    iPad~\cite{ipad} & 2025 & 4V & \checkmark & 98.6 & 98.3 & 88.0 & 94.9 & \textbf{100.0} & 91.7 & -- \\
    SparseDriveV2~\cite{sparsedrivev2} & 2026 & 3V & \checkmark & 98.5 & 98.4 & 88.6 & 95.0 & 99.9 & 92.0 & -- \\
    DriveSuprim~\cite{drivesuprim} & 2026 & 3V & \checkmark & 98.6 & 98.6 & \textbf{91.3} & 95.5 & \textbf{100.0} & 93.5 & 80.0 \\
    RAP-DINO~\cite{rap} & 2026 & 4V & \checkmark & 99.1 & 98.9 & 90.3 & 96.7 & \textbf{100.0} & 93.8 & -- \\
    \midrule
    \ours{} w/o scorer & -- & 7V & $\times$ & \textbf{99.3} & 97.6 & 84.5 & 96.7 & \textbf{100.0} & 90.7 & 102.3 \\
    \ours{} w/ scorer & -- & 7V & \checkmark & 99.1 & \textbf{99.9} & 90.3 & 96.2 & \textbf{100.0} & \textbf{94.2} & 198.4 \\
    \bottomrule
  \end{tabular*}
\end{table*}

We evaluate open-loop planning on NAVSIM v1 \texttt{navtest}~\cite{navsim} (trained on \texttt{navtrain}) via PDMS, and closed-loop driving on the 220 CARLA routes of Bench2Drive~\cite{bench2drive} (trained on the official base split under the standard protocol with trajectory-tracking control~\cite{transfuser}) via Driving Score (DS) and Success Rate (SR).
Both use $H=8$ observed latent slots at $\fv=1$\,Hz and $\Hb=2$ future target slots. Multi-view Cosmos-Predict2.5 uses eight latent frames per view~\cite{cosmos_predict25}, so NAVSIM extends the default 4-frame history to 7\,s using earlier frames from public OpenScene logs~\cite{navsim}.
Both share the same world-action architecture initialized from our pretrained backbone (\cref{fig:pretraining-transfer}a) with a frozen carrier (\TAEW{}~\cite{taehv} on NAVSIM, \DINOvthree{}~\cite{dinov3} on Bench2Drive).
On NAVSIM, the scorer (\cref{sec:method-scorer}) ranks $K=53$ candidates; on 10\,Hz Bench2Drive, to match on-vehicle latency budgets, we omit it ($K=1$) and average over three seeds (\cref{sec:supp-navsim-protocol,sec:supp-b2d-protocol}).

\Cref{tab:timing} instantiates the schedule of \cref{sec:method-formulation}, with a single model per benchmark trained across all scheduled context ages. NAVSIM emulates each age by encoding history ending at $s=t-a_t$; the NAVSIM comparison reports the chunk-step age (1.0\,s), while the same model reaches 94.0 PDMS at the freshness-step age (1.5\,s).
The Video DiT runs on one H20 GPU while the carrier and Action DiT run on a second. Bench2Drive's standard protocol waits for each foreground decision; background world updates advance in simulator time. We additionally inject command delay (\cref{tab:latency-injection}).
We also test open-loop transfer on Waymo~\cite{waymoe2e} and nuScenes~\cite{nuscenes}, reaching 0.54\,m L2 and 0.03\% collision on nuScenes (\cref{sec:supp-nuscenes-transfer,sec:supp-information-flow}).

\subsubsection{Pretraining}
\label{sec:supp-pretraining}
All four stages in \cref{fig:pretraining-transfer}a are fine-tuned on \texttt{navtrain} and evaluated at the chunk-step age without scoring ($K=1$, $\NODE=1$): \texttt{Scratch} initializes the Video DiT randomly, \texttt{Cosmos} initializes from public Cosmos-Predict2.5 weights~\cite{cosmos_predict25}, and $+1\text{M}$ and $+10\text{M}$ add joint video-action pretraining on internal driving clips disjoint from NAVSIM prior to \texttt{navtrain} adaptation. The \texttt{Cosmos} generator underpins the 93.3 PDMS scored system in \cref{sec:exp-sota}, while its frozen-backbone counterpart is analyzed in \cref{sec:supp-generative}. The $+1\text{M}$ and $+10\text{M}$ stages differ solely in pretraining data scale.

\subsubsection{Training settings}
\label{sec:supp-training-settings}
The NAVSIM generator is trained for 10 epochs using AdamW with batch size 64 and learning rates of $10^{-5}$ for the Video DiT and $5\times10^{-4}$ for the Action DiT and freshness adapter; Bench2Drive uses the identical optimizer, batch size, learning rates, and training budget. The trajectory scorer is trained for 10 epochs using AdamW at a learning rate of $10^{-4}$ and batch size 64. Flow noise levels follow the shifted logit-normal distribution of Cosmos-Predict2.5~\cite{cosmos_predict25}, and the joint objective uses loss weights $(\lambda_{\mathrm v},\lambda_{\mathrm a},\lambda_{\mathrm{traj}})=(1,10,3)$.

\begin{figure*}[t]
  \centering
  \includegraphics[width=0.92\textwidth]{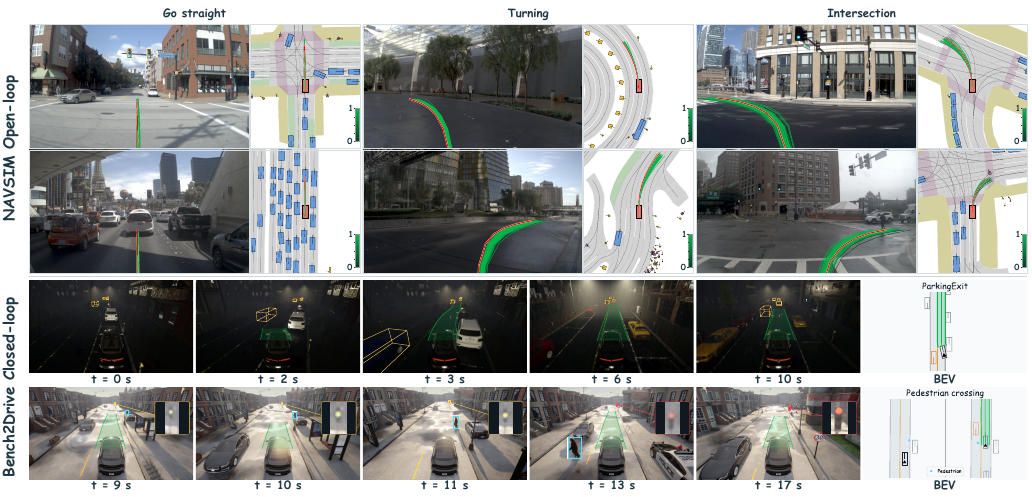}
  \caption{\textbf{Qualitative results.} The top two rows show NAVSIM plans; brighter green indicates higher scorer output and red denotes ground truth. The bottom two rows show closed-loop Bench2Drive examples of rainy-night parking and a pedestrian crossing. Boxes and traffic-light insets are simulator ground-truth overlays.}
  \label{fig:qualitative-results}
\end{figure*}

\subsubsection{NAVSIM protocol}
\label{sec:supp-navsim-protocol}
NAVSIM uses seven $240\times416$ camera views, $H=8$ single-frame observed slots sampled at $\fv=1$\,Hz from 2\,Hz OpenScene logs~\cite{navsim} (spanning 7\,s to match the eight-frame latent temporal capacity of multi-view Cosmos-Predict2.5~\cite{cosmos_predict25}), and $\Hb=2$ future target slots without temporal compression. Its 8D ego condition comprises 2D velocity, 2D acceleration, and a four-way navigation command, and decisions at $\fa=2$\,Hz predict eight waypoints over 4\,s. Training uniformly mixes both scheduled context ages ($a_t\in\{1.0,1.5\}$\,s, corresponding to two or three decisions before $t$) under the schedule in \cref{tab:timing} while $\CurImgs$ and $e_t$ remain current, and the NAVSIM comparison reports the chunk-step age ($1.0$\,s).
The $\Nvoc=21$ trajectory priors are k-means cluster centers of \texttt{navtrain} trajectories mapped into increment space via $G^{-1}$, and normalization scales $\gamma$ are computed strictly from \texttt{navtrain}. Each candidate has $8\times3$ coordinates and an 8D ego condition; the scorer reads $\Nw=21{,}840$ world residual tokens and $\Nf=2{,}730$ freshness tokens.
At evaluation, NAVSIM independently encodes each scene's supplied history clip, including across context-age sweeps.

\subsubsection{Bench2Drive protocol}
\label{sec:supp-b2d-protocol}
Bench2Drive uses six $288\times512$ views, where the causal \WanVAE{}~\cite{wan} encodes a 29-frame clip at 4\,Hz as the initial frame plus seven four-frame groups to produce eight observed slots at 1\,Hz, supplemented by two future target slots during training. Its ego condition combines the target goal with speed, yaw rate, longitudinal and lateral acceleration, pitch, roll, and yaw acceleration, and decisions at 10\,Hz predict 40 waypoints over 4\,s.
Bench2Drive follows the schedule in \cref{tab:timing} and is trained across all ten context ages of its cycle. Standard Bench2Drive evaluation advances CARLA after each foreground decision. Background world updates advance in simulator time, so each cache is first read $\taud$ after its snapshot. The delay-injected evaluation (\cref{sec:supp-delay}) additionally advances CARLA during foreground control latency.
At evaluation, Bench2Drive maintains a persistent streaming cache: once cold start fills the initial $H$ slots, each background update evicts the oldest slot and appends the newest, meaning retained K/V features can carry context propagated from since-evicted frames.

\subsubsection{Cross-dataset implementation}
\label{sec:supp-cross-dataset-impl}
\paragraph{Waymo.}
For Waymo E2E~\cite{waymoe2e} transfer, \ours{} (zero-shot) uses the pretrained weights directly, while \ours{} (fine-tuned) trains an 8-view 480p generator with supervised world-action learning; both draw eight one-step candidates ($K=8$, $\NODE=1$) at zero context age to produce 20 waypoints at 4\,Hz on public validation and hidden test.
When camera history is insufficient, a shared short-history path extrapolates observed ego velocity identically across all configurations and paired conditions.
For nonzero context ages, the generator is paired with the \TAEW{} carrier encoding a five-frame fresh clip ending at $t$; the fresh-input training ablations share a three-epoch fine-tuning budget and 29-frame history from six $480\times832$ cameras paired with five-frame fresh clips, where eight-view freshness adds two physical side cameras.

\paragraph{nuScenes.}
The nuScenes~\cite{nuscenes} transfer and information-flow evaluations use 6,019 validation windows from 150 scenes with six cameras, ego state, and coarse navigation.
In transfer evaluation, \ours{} (zero-shot) uses the pretrained weights, while \ours{} (fine-tuned) uses a fine-tuned generator and a visual trajectory scorer supervised by training-set trajectory-quality targets; both rank $K=53$ candidates (21 priors and 32 noise samples).
In the information-flow study, all three variants share initialization, a three-epoch training budget, and the 6,019 validation windows, generating 53 candidates per window (21 priors with two ODE steps and 32 noise samples with five steps, totaling seven batched action-network calls) ranked by a shared frozen 987M visual scorer.
\subsection{Comparison with published methods}
\label{sec:exp-sota}

\subsubsection{NAVSIM}
In \cref{tab:navsim}, \ours{} achieves \textbf{94.2} PDMS on NAVSIM \texttt{navtest} (top DAC of \textbf{99.9}), surpassing all compared world-action methods (best prior: SimWAM~\cite{simwam}, 91.9) and annotation-supervised planners such as RAP-DINO~\cite{rap} (93.8).
Crucially, this gain stems from the synergy between our asynchronous generator and its coupled 5.2M scorer~\cite{gtrs,drivesuprim} rather than standalone rescoring: from predictive world context and live freshness features, the generator forms a high-ceiling 53-candidate pool (\textbf{98.5} Oracle@53 PDMS vs.\ 90.7 at $K=1$), which the scorer realizes into 94.2 PDMS (\cref{fig:qualitative-results}, top) by reading the generator's world residuals $X_s$ and carrier features $\Zf$ directly (+96\,ms, no separate visual backbone).
The generator requires zero perception labels, learning purely from future video and logged trajectories, whereas only the scorer uses annotation-derived simulator sub-scores (\texttt{Ann.}\ in \cref{tab:navsim}) to align selection with PDMS.
Initialized from public Cosmos-Predict2.5~\cite{cosmos_predict25} weights and trained solely on \texttt{navtrain}, the same architecture still reaches \textbf{93.3} PDMS (88.5 without scoring), exceeding all published world-action methods in \cref{tab:navsim} (\cref{sec:supp-pretraining,sec:supp-generative}).

\FloatBarrier
\subsubsection{Bench2Drive}
\begin{table}[!htbp]
  \caption{\textbf{Bench2Drive comparison.} Camera-only world-action methods on 220 routes under the standard protocol. Emergency Brake and Give Way (\%) assess responses to current events. \ours{}: mean $\pm$ sample std over three seeds of one model; bold: best mean; dashes: unreported. Injected delay: \cref{tab:latency-injection}.}
  \label{tab:b2d}
  \centering
  \footnotesize
  \setlength{\tabcolsep}{2pt}
  \begin{tabular}{@{}lccccc@{}}
    \toprule
    Method & DS $\uparrow$ & SR (\%) $\uparrow$ & \shortstack{Emerg.\\Brake $\uparrow$} & \shortstack{Give\\Way $\uparrow$} & \shortstack{Latency\\(ms) $\downarrow$} \\
    \midrule
    MindDriver~\cite{minddriver} & 65.48 & 39.55 & -- & -- & 1000.0 \\
    Raw2Drive~\cite{raw2drive} & 71.36 & 50.24 & 60.00 & 50.00 & 604.0 \\
    World4Scorer~\cite{world4scorer} & 73.27 & 46.36 & -- & -- & -- \\
    MomWorld~\cite{momworld} & 74.07 & 50.00 & 61.71 & 50.00 & -- \\
    GraphWorld~\cite{graphworld} & 76.71 & 44.22 & 49.09 & 50.00 & -- \\
    PerlAD~\cite{perlad} & 78.70 & \textbf{57.27} & 68.33 & 50.00 & -- \\
    WPT (teacher)~\cite{wpt} & 79.23 & 54.54 & 65.00 & 50.00 & -- \\
    UniDrive-WM~\cite{unidrivewm} & 79.31 & 56.42 & 79.98 & 40.00 & 2500.0 \\
    \midrule
    \ours{} & \textbf{80.30}$\pm0.32$ & 56.36$\pm0.79$ & \textbf{88.33} & \textbf{60.00} & 109.9 \\
    \bottomrule
  \end{tabular}
\end{table}

On Bench2Drive (\cref{tab:b2d}), \ours{} achieves \textbf{80.30} DS and 56.36\% SR over the 220 routes, the highest DS among compared world-action methods, while combining world context aged 1.0--1.9\,s with current visual evidence at every decision.
Foreground decisions take 109.9\,ms, compared with 604--2500\,ms for the other methods reporting runtime (\cref{sec:supp-comparator-timing}).
With one world update per ten decisions and no scorer, \ours{} reaches \textbf{88.33\%} on Emergency Brake and \textbf{60.00\%} on Give Way in multi-ability evaluation~\cite{bench2drive} (\cref{fig:qualitative-results}, bottom); masking $\mathcal F_{t|s}$ with fixed weights drops DS by 10.8 points, SR by 16.8 points, and these abilities to 61.67\% and 40.00\%.

\FloatBarrier
\subsection{Cross-dataset transfer}
\label{sec:supp-native}
\begin{table}[!htbp]
\centering\footnotesize
\setlength{\tabcolsep}{3pt}
\caption{\textbf{Waymo E2E comparison.} Representative methods on public validation and hidden test (RFS $\uparrow$). Top: zero-shot transfer. Bold: \ours{}; RL: reinforcement learning.}
\label{tab:transfer-waymo}
\begin{tabular*}{\columnwidth}{@{\extracolsep{\fill}}lrr@{}}\toprule
Method & Public val. & Hidden test \\
\midrule
Qwen2.5-VL 7B~\cite{poutine} & 5.40 & -- \\
Qwen2.5-VL 3B~\cite{poutine} & 5.59 & -- \\
Poutine-Base (CoVLA)~\cite{poutine} & 7.74 & -- \\
\textbf{\ours{} (zero-shot)} & \textbf{7.7841} & -- \\
\midrule
\multicolumn{3}{@{}l}{\emph{Other end-to-end planners}} \\
RAP-DINO~\cite{rap,mindvlau1} & 7.91 & -- \\
MindVLA-U1 + MoT~\cite{mindvlau1} & 7.92 & -- \\
Qwen-Drive-1.0-SFT~\cite{qwendrive} & 7.95 & -- \\
Streaming Intent~\cite{streamingintent} & 7.96 & -- \\
Poutine-Base~\cite{poutine} & 8.12 & 7.91 \\
AutoVLA~\cite{autovla} & -- & 7.5566 \\
IRL-VLA~\cite{irlvla,ntr} & -- & 7.8900 \\
Poutine~\cite{poutine,autovla} & -- & 7.9860 \\
\midrule
\multicolumn{3}{@{}l}{\emph{World-action models}} \\
SimWAM~\cite{simwam,simwamrepo} & 7.99 & 7.77 \\
SimWAM + RL~\cite{simwamrepo} & 8.29 & 7.84 \\
SUV~\cite{suv} & -- & 7.94 \\
\textbf{\ours{} (fine-tuned)} & \textbf{8.2723} & \textbf{8.0050} \\
\bottomrule\end{tabular*}
\end{table}

\subsubsection{Waymo}
Across all 478 public-validation scenes (479 decisions), \ours{} (zero-shot) and \ours{} (fine-tuned) attain \textbf{7.7841} and \textbf{8.2723} Rater Feedback Score (RFS; \cref{tab:transfer-waymo}).
On hidden test, \ours{} (fine-tuned) reaches \textbf{8.0050}, leading reported world-action models ahead of SUV (7.94) and SimWAM + RL (7.84)~\cite{suv,simwamrepo} (\cref{sec:supp-cross-dataset-impl,sec:supp-transfer}).

\subsubsection{Waymo transfer results}
\label{sec:supp-transfer}
The transfer results in \cref{tab:transfer-waymo,tab:waymo-details} follow the Waymo protocol in \cref{sec:supp-cross-dataset-impl}. The freshness comparison in \cref{sec:supp-freshness} holds single-trajectory inference ($K=1$), model weights, and paired inputs fixed across the on/off conditions.

\paragraph{Complete public validation.}
All public results cover 478 scenes and 479 scored decisions. \ours{} (zero-shot) and \ours{} (fine-tuned) attain \textbf{7.7841} and \textbf{8.2723} RFS, respectively (\cref{tab:transfer-waymo,tab:waymo-details}). RFS averages ten category means; displacement errors average all 479 decisions.
\begin{table}[!htbp]
\centering\footnotesize
\setlength{\tabcolsep}{2.5pt}
\caption{\textbf{Waymo public-validation transfer results.} Both systems use eight one-step candidates ($K=8$, $\NODE=1$) at $a_t=0$ across all 478 scenes (479 decisions). Errors are in meters.}
\label{tab:waymo-details}
\begin{tabular*}{\columnwidth}{@{\extracolsep{\fill}}lrrrr@{}}\toprule
Model & RFS $\uparrow$ & ADE$_3$ & ADE$_5$ & FDE$_5$ \\
\midrule
\ours{} (zero-shot) & 7.7841 & 1.3353 & 2.8773 & 7.1533 \\
\ours{} (fine-tuned) & 8.2723 & 1.0826 & 2.3365 & 6.0575 \\
\bottomrule\end{tabular*}
\end{table}

\paragraph{Hidden-test performance.}
Submitted \ours{} (fine-tuned) achieves \textbf{8.005038} RFS, with ADE$_3$ of 1.2094878\,m and ADE$_5$ of 2.5969467\,m (\cref{tab:waymo-hidden}). Its eleven category scores average to 8.00503843, matching the official overall score after rounding.
This is the highest publicly reported hidden-test RFS among world-action models. It exceeds SUV's reported 7.94~\cite{suv} by approximately \textbf{0.0650}; SimWAM reports 7.77 after fine-tuning and 7.84 with reinforcement learning~\cite{simwamrepo}, giving \ours{} a \textbf{0.165038} advantage over the latter.
\begin{table}[!htbp]
\centering\footnotesize
\setlength{\tabcolsep}{4pt}
\caption{\textbf{Waymo hidden-test comparison.} Representative methods with published precision retained. ADE is in meters. Bold: \ours{}.}
\label{tab:waymo-hidden-comparison}
\begin{tabular*}{\columnwidth}{@{\extracolsep{\fill}}lrrr@{}}\toprule
Method & RFS $\uparrow$ & ADE$_3$ $\downarrow$ & ADE$_5$ $\downarrow$ \\
\midrule
\multicolumn{4}{@{}l}{\emph{Vision-language-action planners}} \\
AutoVLA~\cite{autovla} & 7.5566 & 1.3507 & 2.9580 \\
IRL-VLA~\cite{irlvla,ntr} & 7.8900 & 1.2184 & 2.8235 \\
Poutine-Base~\cite{poutine} & 7.91 & 1.27 & 2.94 \\
Poutine~\cite{poutine,autovla} & 7.9860 & 1.2055 & 2.7419 \\
\midrule
\multicolumn{4}{@{}l}{\emph{World-action models}} \\
SimWAM~\cite{simwamrepo} & 7.77 & 1.22 & 2.69 \\
SimWAM + RL~\cite{simwamrepo} & 7.84 & 1.21 & 2.68 \\
SUV~\cite{suv} & 7.94 & 1.24 & 2.90 \\
\textbf{\ours{} (fine-tuned)} & \textbf{8.0050} & \textbf{1.2095} & \textbf{2.5969} \\
\bottomrule\end{tabular*}
\end{table}

\begin{table}[!htbp]
\centering\footnotesize
\setlength{\tabcolsep}{5pt}
\caption{\textbf{Official Waymo hidden-test results of \ours{} (fine-tuned).} RFS is the equal-weight mean over all 11 scenario categories. Displacement errors are in meters.}
\label{tab:waymo-hidden}
\begin{tabular*}{\columnwidth}{@{\extracolsep{\fill}}lr@{}}\toprule
Metric / category & Value \\
\midrule
\textbf{Overall RFS} & \textbf{8.005038} \\
\midrule
Construction & 8.446138 \\
Intersection & 8.112481 \\
Pedestrian & 8.085993 \\
Cyclist & 7.9328837 \\
Multi Lane Maneuver & 7.914063 \\
Single Lane Maneuver & 8.452082 \\
Cut In & 8.227605 \\
Foreign Object Debris & 8.13 \\
Special Vehicle & 8.14193 \\
Spotlight & 7.2074757 \\
Others & 7.4047713 \\
\midrule
ADE$_3$ $\downarrow$ & 1.2094878 \\
ADE$_5$ $\downarrow$ & 2.5969467 \\
\bottomrule\end{tabular*}
\end{table}

\paragraph{Published comparisons.}
Public-validation comparisons in \cref{tab:transfer-waymo} follow Poutine Table~1 (arXiv v1)~\cite{poutine}, MindVLA-U1 Table~1~\cite{mindvlau1}, Qwen-Drive Table~4(a)~\cite{qwendrive}, and SimWAM's official release~\cite{simwamrepo}; Streaming Intent~\cite{streamingintent} reports on 438 sequences.
For hidden test (\cref{tab:waymo-hidden-comparison}), Poutine and AutoVLA use the leaderboard values in AutoVLA Table~S3~\cite{autovla}; Poutine-Base uses Poutine Table~2 (arXiv v4)~\cite{poutine}; IRL-VLA~\cite{irlvla} follows the leaderboard entry reported in NTR Table~1~\cite{ntr}; SUV uses its Table~3~\cite{suv}; and SimWAM uses its official release~\cite{simwamrepo}. Poutine-Base and Poutine are separate published configurations. SUV and SimWAM jointly learn visual dynamics and actions, while the vision-language-action planners provide complementary comparisons.

\begin{table*}[t]
\centering\footnotesize
\setlength{\tabcolsep}{3pt}
\caption{\textbf{nuScenes open-loop planning comparison.} Published rows follow the SimWAM compilation~\cite{simwam,simwamrepo}. Camera$^*$ uses the front view; Aux. denotes dense perception prediction tasks. \ours{} rows use 6,019 windows and average the 1, 2, and 3\,s endpoints. Bold marks the best values at the displayed precision.}
\label{tab:nuscenes-comparison}
\begin{tabular*}{\textwidth}{@{\extracolsep{\fill}}lcccrrrrrrrr@{}}\toprule
\multirow{2}{*}{Method} & \multirow{2}{*}{Fine-tuned} & \multirow{2}{*}{Input} & \multirow{2}{*}{Aux.} & \multicolumn{4}{c}{L2 (m) $\downarrow$} & \multicolumn{4}{c}{Collision rate (\%) $\downarrow$} \\
\cmidrule(lr){5-8}\cmidrule(lr){9-12}
 & & & & 1\,s & 2\,s & 3\,s & Avg. & 1\,s & 2\,s & 3\,s & Avg. \\
\midrule
ST-P3 & Yes & Camera & Map\&Box\&Depth & 1.33 & 2.11 & 2.90 & 2.11 & 0.23 & 0.62 & 1.27 & 0.71 \\
UniAD & Yes & Camera & Map\&Box\&Motion & 0.48 & 0.96 & 1.65 & 1.03 & 0.05 & 0.17 & 0.71 & 0.31 \\
VAD-Base & Yes & Camera & Map\&Box\&Motion & 0.54 & 1.15 & 1.98 & 1.22 & 0.04 & 0.39 & 1.17 & 0.53 \\
GenAD & Yes & Camera & Map\&Box\&Motion & 0.36 & 0.83 & 1.55 & 0.91 & 0.06 & 0.23 & 1.00 & 0.43 \\
Epona & Yes & Camera$^*$ & None & 0.61 & 1.17 & 1.98 & 1.25 & 0.01 & 0.22 & 0.85 & 0.36 \\
DriveVA & No & Camera$^*$ & None & 0.33 & 0.76 & 1.43 & 0.84 & \textbf{0.00} & 0.07 & 0.12 & 0.06 \\
DriveWAM & No & Camera$^*$ & None & 0.28 & 0.81 & 1.80 & 0.96 & \textbf{0.00} & 0.05 & 0.14 & 0.06 \\
SimWAM & No & Camera$^*$ & None & 0.29 & 0.82 & 1.77 & 0.96 & \textbf{0.00} & 0.03 & 0.11 & 0.05 \\
\midrule
\textbf{\ours{} (zero-shot)} & No & Camera & None & 0.30 & 0.77 & 1.61 & 0.89 & \textbf{0.00} & 0.05 & 0.10 & 0.05 \\
\textbf{\ours{} (fine-tuned)} & Yes & Camera & None & \textbf{0.18} & \textbf{0.47} & \textbf{0.98} & \textbf{0.54} & 0.02 & \textbf{0.02} & \textbf{0.05} & \textbf{0.03} \\
\bottomrule\end{tabular*}
\end{table*}

\subsubsection{nuScenes transfer results}
\label{sec:supp-nuscenes-transfer}
\Cref{tab:nuscenes-comparison} evaluates \ours{} (zero-shot) and \ours{} (fine-tuned) on the 6,019 nuScenes validation windows under the $K=53$ visual-scoring protocol (\cref{sec:supp-cross-dataset-impl}). \ours{} (zero-shot) achieves 0.8943\,m mean endpoint L2 and 0.0498\% mean collision rate; \ours{} (fine-tuned) reaches \textbf{0.5434}\,m and \textbf{0.0277\%}, respectively.

In \cref{tab:nuscenes-comparison}, \texttt{Aux.} denotes dense map, box, depth, or motion prediction tasks. Our generators learn from future video and ego trajectories; the visual scorers learn ranking targets from training-set planning outcomes.
L2 and collision rate are averaged over the 1, 2, and 3\,s endpoints, as defined in \cref{sec:supp-information-flow}. Published comparator values follow SimWAM~\cite{simwam,simwamrepo}.

\subsection{Ablations and analysis}
\label{sec:exp-ablation}
\label{sec:supp-exp}
\begin{figure}[!htbp]
  \centering
  \includegraphics[width=\columnwidth]{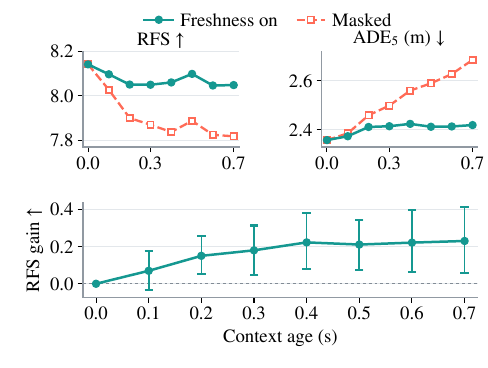}
  \caption{\textbf{Freshness under delayed world context.} \ours{} (fine-tuned) RFS and ADE$_5$ with freshness enabled or masked (top); paired RFS gain and pointwise 95\% scene-bootstrap intervals (bottom). Both conditions coincide at $a_t=0$.}
  \label{fig:freshness-native}
\end{figure}

\subsubsection{Freshness reliance under delayed context}
Masking $\mathcal F_{t|s}$ at inference quantifies reliance on live images as world context ages.
On Waymo public validation ($a_t \in [0, 0.7]$\,s set offline with \TAEW{}, $K=1$, $\NODE=1$), freshness raises category-averaged RFS by \textbf{0.18} across nonzero ages (95\% CI $[0.08,0.30]$; \cref{fig:freshness-native}; \cref{sec:supp-freshness,sec:supp-fresh-input}).
On NAVSIM, masking \TAEW{} costs 0.1 PDMS at zero age vs.\ 1.7 and 3.1 at 1.0\,s and 1.5\,s ages, showing live feedback sustains planning as world context ages (\cref{sec:supp-navsim-age}).

\subsubsection{Freshness pairing and uncertainty}
\label{sec:supp-freshness}
Holding decision timestamp $t$ fixed under the Waymo setup of \cref{sec:supp-cross-dataset-impl}, we shift the clean-history endpoint $s$ offline to sweep context ages $a_t=t-s\in\{0,0.1,\ldots,0.7\}$\,s, measuring the effect of context age at a fixed decision time. Each paired evaluation covers all 478 scenes and 479 scored decisions, with the same model weights, ego states, references, decision identities, and Gaussian seeds across ages and conditions. Each evaluation directly outputs a single one-step trajectory ($K=1$, $\NODE=1$) of 20 finite waypoints: at nonzero ages, masking removes $\mathcal F_{t|s}$ while retaining $(\mathcal C_s,e_t)$, whereas at zero age the carrier is bypassed so both conditions share an identical \emph{boundary prediction}.

Waymo's RFS macro-averages human-preference scores across ten scenario categories, while ADE$_5$ and FDE$_5$ measure displacement against the highest-rated human reference over 20 waypoints at 4\,Hz (5\,s horizon). Confidence intervals use 2,000 paired, category-stratified scene-bootstrap draws that share each scene resample across all ages and conditions; the \emph{mean-seven} statistic averages nonzero-age contrasts within each draw, and per-age intervals are pointwise.
\ours{} (fine-tuned) uses 8-view 480p freshness; its weights are fixed across all paired conditions, with a boundary RFS of 8.1418.

\begin{table}[!htbp]
\centering\footnotesize
\setlength{\tabcolsep}{1.5pt}
\caption{\textbf{Paired Waymo freshness results.} Both models use $K=1$ on all 478 scenes (479 decisions per age). Paired columns show on/off values; ADE$_5$ is in meters.}
\label{tab:fresh-ft-values}
\begin{tabular*}{\columnwidth}{@{\extracolsep{\fill}}rccrl@{}}\toprule
$a_t$ (s) & RFS on/off & ADE$_5$ on/off & RFS gain & 95\% interval \\
\midrule
\multicolumn{5}{@{}l}{\emph{\ours{} (fine-tuned)}} \\
0.0 & 8.1418/8.1418 & 2.3582/2.3582 & 0.0000 & [0.0000, 0.0000] \\
0.1 & 8.0972/8.0273 & 2.3740/2.3859 & 0.0699 & [-0.0324, 0.1780] \\
0.2 & 8.0509/7.9004 & 2.4117/2.4595 & 0.1505 & [0.0512, 0.2593] \\
0.3 & 8.0502/7.8700 & 2.4148/2.4990 & 0.1802 & [0.0471, 0.3139] \\
0.4 & 8.0606/7.8376 & 2.4244/2.5601 & 0.2230 & [0.0814, 0.3798] \\
0.5 & 8.0990/7.8877 & 2.4126/2.5901 & 0.2113 & [0.0748, 0.3443] \\
0.6 & 8.0468/7.8251 & 2.4139/2.6276 & 0.2218 & [0.0640, 0.3972] \\
0.7 & 8.0487/7.8181 & 2.4195/2.6855 & 0.2306 & [0.0562, 0.4155] \\
\midrule
\multicolumn{5}{@{}l}{\emph{\ours{} (zero-shot)}} \\
0.0 & 7.5561/7.5561 & 3.2686/3.2686 & 0.0000 & [0.0000, 0.0000] \\
0.1 & 7.5036/7.4846 & 3.1844/3.2482 & 0.0190 & [-0.0549, 0.0789] \\
0.2 & 7.5698/7.5533 & 3.1734/3.2516 & 0.0165 & [-0.0615, 0.0840] \\
0.3 & 7.5120/7.5157 & 3.1414/3.2219 & -0.0037 & [-0.0899, 0.0714] \\
0.4 & 7.5502/7.5207 & 3.1336/3.2196 & 0.0295 & [-0.0552, 0.1148] \\
0.5 & 7.5601/7.4584 & 3.1268/3.2241 & 0.1017 & [-0.0139, 0.2006] \\
0.6 & 7.5416/7.4167 & 3.1124/3.2258 & 0.1249 & [0.0346, 0.2158] \\
0.7 & 7.5446/7.3492 & 3.1098/3.2336 & 0.1954 & [0.0995, 0.2931] \\
\bottomrule\end{tabular*}
\end{table}

Across the seven nonzero ages, enabling freshness yields a mean-seven RFS gain of 0.0691 (95\% CI $[0.0016,0.1310]$) for \ours{} (zero-shot), reaching $+0.1954$ ($[0.0995,0.2931]$) at $0.7$\,s. For \ours{} (fine-tuned), it raises mean-seven RFS from 7.8809 (masked) to 8.0648 (enabled), a paired gain of 0.1839 ($[0.0810,0.2990]$). Per-age values and category breakdowns appear in \cref{tab:fresh-ft-values,tab:fresh-categories}.
\begin{table}[!htbp]
\centering\footnotesize
\setlength{\tabcolsep}{3pt}
\caption{\textbf{\ours{} (fine-tuned) category breakdown.} Mean over seven nonzero ages; $n$ denotes decisions per age, totaling 479.}
\label{tab:fresh-categories}
\begin{tabular*}{\columnwidth}{@{\extracolsep{\fill}}lrrrr@{}}\toprule
Category & $n$ & RFS on & RFS off & Gain \\
\midrule
Construction & 15 & 8.3648 & 8.3668 & -0.0020 \\
Cut ins & 20 & 8.6655 & 8.2505 & 0.4150 \\
Cyclist & 71 & 8.3676 & 8.1399 & 0.2277 \\
Foreign Object Debris & 78 & 8.0461 & 7.8824 & 0.1638 \\
Intersections & 116 & 7.8786 & 7.8275 & 0.0510 \\
Multi-Lane Maneuvers & 42 & 7.5356 & 7.5341 & 0.0014 \\
Others & 22 & 7.7122 & 7.4562 & 0.2560 \\
Pedestrian & 52 & 8.3192 & 8.2007 & 0.1185 \\
Single-Lane Maneuvers & 38 & 8.4640 & 8.1514 & 0.3126 \\
Special Vehicles & 25 & 7.2941 & 6.9992 & 0.2949 \\
\bottomrule\end{tabular*}
\end{table}

\subsubsection{Context age on NAVSIM}
\label{sec:supp-navsim-age}
Using a single \TAEW{} generator trained on the scheduled 1.0\,s and 1.5\,s ages ($\Tpred=2.0$\,s), \cref{tab:navsim-context-age} evaluates five context ages ($K=1$, $\NODE=1$, no scorer), where $0.0$, $0.5$, and $2.0$\,s lie outside the training ages. Each condition independently encodes eight 1\,Hz history slots ending at $s=t-a_t$, holding decision timestamps, current images, ego states, scene population, and seeds fixed across all ages and on/off pairs. Freshness is enabled or masked at every age, including zero. With freshness enabled, PDMS stays within 0.2 points up to 1.5\,s before dropping to 89.4 at 2.0\,s, whereas paired reliance grows steadily across ages.
\begin{table}[!htbp]
  \caption{\textbf{NAVSIM context age with the \TAEW{} generator.} Reliance is the paired freshness-on minus masked PDMS.}
  \label{tab:navsim-context-age}
  \centering
  \footnotesize
  \setlength{\tabcolsep}{0pt}
  \begin{tabular*}{\columnwidth}{@{\extracolsep{\fill}}lccc@{}}
    \toprule
    Context age $a_t$ (s) & Freshness on & Masked & Reliance \\
    \midrule
    0.0 & 90.9 & 90.8 & 0.1 \\
    0.5 & 90.8 & 90.0 & 0.8 \\
    1.0 (chunk) & 90.7 & 89.0 & 1.7 \\
    1.5 (freshness) & 90.9 & 87.8 & 3.1 \\
    2.0 & 89.4 & 83.8 & 5.6 \\
    \bottomrule
  \end{tabular*}
\end{table}

\subsubsection{Current-observation encoders}

At the NAVSIM chunk-step age ($K=1$, $\NODE=1$, no scorer), the reconstruction-based \TAEW{} carrier reaches 90.7 PDMS and \DINOvthree{} reaches 90.6; masking live visual evidence lowers each by 1.7 points, showing effective visual feedback from compact reconstruction features (\cref{sec:supp-exp-age}).

\subsubsection{Current-observation encoders and freshness carriers}
\label{sec:supp-exp-age}
\Cref{tab:carriers} compares carriers on seven-view NAVSIM \texttt{navtest} at the chunk-step age $a_t=1.0$\,s, using one Gaussian candidate and one Euler step ($K=1$, $\NODE=1$) without scoring. The reconstruction-based \TAEW{} and semantic \DINOvthree{} carriers reach 90.7 and 90.6 PDMS, respectively; each loses 1.7 points when its freshness bank is masked with the model weights fixed. These results establish effective current-observation feedback with a reconstruction-based encoder; an additional semantic backbone is not required for this interface.
Here, \emph{Pass} indicates whether current images invoke the Video DiT: lightweight carriers require no Video DiT call, whereas native \WanVAE{} executes a shared, stop-gradient Video DiT pass over a nine-frame clip. The ratio $\Fcar/\Fw$ compares carrier FLOPs against a full-history world pass (excluding the native 3D VAE). Carrier latency is measured on H20 over seven cameras; native \WanVAE{} includes VAE encoding and its shared Video DiT pass.
We distinguish two ways of removing freshness: (i) \emph{inference-time masking} removes only $\mathcal F_{t|s}$ while holding weights, world context, scenes, $e_t$, and Gaussian seeds fixed (\cref{sec:exp-ablation,tab:navsim-context-age}); and (ii) \emph{architectural removal} eliminates the carrier and retrains synchronous execution at $a_t=0$, where current images reach the action model through the world branch (\cref{sec:supp-sync}).
\begin{table}[!htbp]
  \caption{\textbf{Freshness carriers at the chunk-step age.} Seven-view NAVSIM, $a_t=1.0$\,s, $K=1$, $\NODE=1$, no scorer; carrier cost on H20. Masked: PDMS with $\mathcal F_{t|s}$ removed at inference; Reliance: their difference.}
  \label{tab:carriers}
  \centering
  \footnotesize
  \setlength{\tabcolsep}{0pt}
  \begin{tabular*}{\columnwidth}{@{\extracolsep{\fill}}lcccccc@{}}
    \toprule
    Model & Pass & $\Fcar/\Fw$ & ms & PDMS $\uparrow$ & Masked & Reliance \\
    \midrule
    \DINOvthree{} & no & 0.34\% & \textbf{16.97} & 90.6 & 88.9 & 1.7 \\
    \TAEW{} & no & 0.30\% & 17.84 & \textbf{90.7} & 89.0 & 1.7 \\
    \TransFuserName{} & no & \textbf{0.13\%} & 19.23 & 90.3 & 88.5 & 1.8 \\
    Native \WanVAE{} & shared & 18.00\% & 1418.51 & 90.6 & 89.0 & 1.6 \\
    \bottomrule
  \end{tabular*}
\end{table}

\subsubsection{Fresh-input configurations}
\label{sec:supp-fresh-input}
\Cref{tab:carrier-native} compares freshness camera configurations on Waymo under the three-epoch fine-tuning protocol in \cref{sec:supp-cross-dataset-impl}. Boundary rows ($a_t=0$) bypass freshness to compare backbone training configurations directly, whereas nonzero-age rows average freshness-enabled RFS and ADE$_5$ over $a_t\in\{0.1,\ldots,0.7\}$\,s.
\begin{table}[!htbp]
\centering\footnotesize
\setlength{\tabcolsep}{3pt}
\caption{\textbf{Fresh-input configurations after three epochs of fine-tuning.} At $a_t=0$ the carrier is bypassed; nonzero-age rows average freshness-enabled results over $a_t=0.1,\ldots,0.7$\,s. All rows cover 478 scenes and 479 decisions per age.}
\label{tab:carrier-native}
\begin{tabular*}{\columnwidth}{@{\extracolsep{\fill}}llrr@{}}\toprule
\shortstack{Fresh training input\\views $\times$ H $\times$ W} & $a_t$ (s) & RFS $\uparrow$ & ADE$_5$ (m) $\downarrow$ \\
\midrule
\multirow{2}{*}{6 $\times$ 480 $\times$ 832} & 0 & 8.0897 & 2.4150 \\
 & 0.1--0.7 & 8.0444 & 2.4213 \\
\multirow{2}{*}{6 $\times$ 240 $\times$ 416} & 0 & 8.0176 & 2.3579 \\
 & 0.1--0.7 & 7.9613 & 2.4603 \\
\multirow{2}{*}{8 $\times$ 480 $\times$ 832} & 0 & 8.1348 & 2.3799 \\
 & 0.1--0.7 & 8.0795 & 2.4231 \\
\bottomrule\end{tabular*}
\end{table}

\subsubsection{Predictive span}

\begin{table}[!htbp]
  \caption{\textbf{Predictive span and context age on NAVSIM.} PDMS without the scorer, with the \TAEW{} carrier. Ages 1.0 and 1.5\,s follow the deployment schedule; 2.0\,s is outside it.}
  \label{tab:predictive-span}
  \centering
  \footnotesize
  \setlength{\tabcolsep}{0pt}
  \begin{tabular*}{\columnwidth}{@{\extracolsep{\fill}}lccc@{}}
    \toprule
    Span $\Tpred$ ($\Hb$) & $a_t{=}1.0$\,s & $a_t{=}1.5$\,s & $a_t{=}2.0$\,s \\
    \midrule
    1.0\,s (1) & 89.9 & 89.2 & 88.6 \\
    2.0\,s (2), \ours{} & 90.7 & 90.9 & 89.4 \\
    4.0\,s (4) & 90.2 & 89.7 & 89.6 \\
    \bottomrule
  \end{tabular*}

\end{table}
\Cref{tab:predictive-span} varies the trained predictive span by retraining with $\Hb\in\{1,2,4\}$ while fixing the 10M-clip initialization, eight-slot history, \TAEW{} carrier, training ages, and execution schedule.
The 2.0\,s default is the shortest span covering $\Tupd+\tauw=1.76$\,s on NAVSIM (\cref{eq:m-age}) and performs best at both scheduled context ages: shortening it to 1.0\,s loses 0.8 and 1.7 PDMS, while extending it to 4.0\,s loses 0.5 and 1.2 points.
The 4.0\,s span is slightly higher only at the out-of-schedule 2.0\,s age; ablating $\mathcal C_s$ ($87.9$ PDMS), $g^\ell=1$, 4-frame history, or single-head scoring also hurts PDMS (\cref{sec:supp-exp-ablation}).

\subsubsection{World context and action component ablations}
\label{sec:supp-exp-ablation}
\begin{table}[!htbp]
\centering\footnotesize
\setlength{\tabcolsep}{3pt}
\caption{\textbf{Design ablations on NAVSIM.} Each row changes one factor relative to its group's reference.}
\label{tab:design-ablation}
\begin{tabular*}{\columnwidth}{@{\extracolsep{\fill}}lr@{}}
\toprule
Variant & PDMS $\uparrow$ \\
\midrule
\multicolumn{2}{@{}l}{Without scorer, $K=1$ and $\NODE=1$} \\
Reference & 90.7 \\
Trained without world context $\mathcal C_s$ & 87.9 \\
Freshness carrier masked (eval.\ only) & 89.0 \\
Fixed value gates $g^\ell=1$ & 90.1 \\
Default 4-frame history, padded (eval.\ only) & 89.3 \\
\midrule
\multicolumn{2}{@{}l}{With scorer, $K=53$ and $\NODE=1$} \\
Six-head reference & 94.2 \\
Single PDMS head & 93.7 \\
\bottomrule
\end{tabular*}

\end{table}
\Cref{tab:design-ablation} ablates core choices on NAVSIM: removing world context $\mathcal C_s$, masking the freshness carrier, fixing freshness value gates to $g^\ell=1$, padding the default 4-frame history instead of using a 7\,s history, or replacing the six sub-score heads with a single PDMS head each lowers PDMS. Removing $\mathcal C_s$, fixing $g^\ell=1$, and using a single PDMS head are retrained variants, whereas freshness masking and 4-frame padded history are evaluation-only probes.

\Cref{tab:supp-ablation} extends \cref{tab:design-ablation}, evaluating unscored variants against the 90.7 PDMS chunk-step reference (fixed seed, $K=1$, $\NODE=1$) and scored variants against the 94.2 PDMS reference ($K=53$). Broadening training-age sampling to $a_t\in\{0.5,\dots,2.0\}$\,s~\cite{ahawam} instead of the default 1.0 and 1.5\,s ages yields 90.6 PDMS.
In \cref{tab:design-ablation}, the default-history probe fills eight slots from NAVSIM's standard four-frame input (1.5\,s of video, unrelated to the 1.5\,s context age) by repeating the earliest frame, which both truncates temporal context and introduces a train--test mismatch; closed-loop deployment instead uses a brief stationary cold start to accumulate eight distinct slots.
Increasing Euler integration steps ($\NODE\in\{2,5\}$) multiplies foreground action cost without improving PDMS and slightly degrades scored accuracy by shifting candidate distributions away from the one-step-trained scorer. Finally, vocabulary-only ($K=21$), Gaussian-only ($K=32$), and hybrid ($K=53$) candidate pools confirm the benefit of combining structured priors with stochastic draws.
\begin{table}[!htbp]
  \caption{\textbf{Additional NAVSIM component ablations.} Each row changes one factor, except candidate pools, whose composition and count both vary.}
  \label{tab:supp-ablation}
  \centering
  \footnotesize
  \setlength{\tabcolsep}{0pt}
  \begin{tabular*}{\columnwidth}{@{\extracolsep{\fill}}llcc@{}}
    \toprule
    Factor & Variant & Ref. & PDMS $\uparrow$ \\
    \midrule
    \multicolumn{4}{@{}l}{\emph{Without scorer} ($K{=}1$, one ODE step)} \\
    Training ages & $a_t\in\{0.5,\dots,2.0\}$\,s & 90.7 & 90.6 \\
    ODE steps & $\NODE{=}2$ / $\NODE{=}5$ & 90.7 & 90.6\,/\,90.4 \\
    \midrule
    \multicolumn{4}{@{}l}{\emph{With scorer} ($K{=}53$)} \\
    Candidates & vocabulary-only pool & 94.2 & 93.5 \\
     & Gaussian-only pool & 94.2 & 93.7 \\
    Candidate count & $K{=}1$ / $K{=}16$ & 94.2 & 90.7\,/\,93.7 \\
    ODE steps & $\NODE{=}2$ / $\NODE{=}5$ & 94.2 & 94.0\,/\,93.6 \\
    Carrier type & native \WanVAE{} & 94.2 & 94.0 \\
    \bottomrule
  \end{tabular*}
\end{table}

\subsubsection{Information-flow evaluation}
\label{sec:supp-information-flow}
On nuScenes~\cite{nuscenes}, we compare planning from single-frame context, clean history, and explicit future-video generation prior to action decoding (\emph{imagine-then-execute}; \cref{tab:flow-native}). Clean history lowers mean endpoint L2 error from 0.5974 to 0.5323\,m at an unchanged mean collision rate of 0.5760\%; imagine-then-execute obtains 1.0717\,m, supporting direct planning from clean history. Per-horizon scores appear in \cref{tab:flow-details}.
\begin{table}[!htbp]
\centering\footnotesize
\setlength{\tabcolsep}{3pt}
\caption{\textbf{Information-flow ablation on nuScenes.} Endpoint error and collision (C) are averaged over 1, 2, and 3\,s; C$_3$ is collision at 3\,s. Lower is better.}
\label{tab:flow-native}
\begin{tabular*}{\columnwidth}{@{\extracolsep{\fill}}lrrr@{}}\toprule
System & Mean L2 (m) & Mean C (\%) & C$_3$ (\%) \\
\midrule
Single-frame context & 0.5974 & 0.5760 & 0.9802 \\
\ours{}: clean history & 0.5323 & 0.5760 & 0.7642 \\
Imagine then execute & 1.0717 & 0.7753 & 1.1962 \\
\bottomrule\end{tabular*}
\end{table}

All three variants share the initialization, training budget, validation windows, and $K=53$ candidate/scorer protocol specified in \cref{sec:supp-cross-dataset-impl}.
Specifically, \emph{single-frame context}, in the style of Fast-WAM~\cite{fastwam}, supplies only the current frame to both generator and scorer, whereas \emph{clean history} and \emph{imagine-then-execute} supply eight history frames to the generator and the final two observed frames to the scorer: clean-history training supervises future video and actions independently for direct test-time action decoding, while imagine-then-execute runs eight video-network passes before predicting actions. The visual inputs of both generator and scorer follow each information-flow setting.

\paragraph{Metric definitions.}
Defining waypoint error $\ell_j=\|\hat{\mathbf x}_{j/2}-\mathbf x_{j/2}\|_2$, mean L2 averages the 1, 2, and 3\,s endpoints $(\ell_2+\ell_4+\ell_6)/3$ across windows (per-horizon values in \cref{tab:flow-details}).
Collision evaluation follows the frozen UniAD occupancy protocol~\cite{uniad} on a $200\times200$ grid (0.5\,m resolution) using an axis-aligned $4.084\times1.85$\,m ego footprint shifted 0.5\,m forward against eight vehicle classes: ground-truth-collided timestamps are excluded from the numerator while retaining the full 6,019-window denominator (with invalid future horizons contributing zero under the all-validation convention; horizon-valid counts at $0.5,\dots,3.0$\,s are 5,869, 5,719, 5,569, 5,419, 5,269, and 5,119), while \cref{tab:flow-complete} restricts the same predictions to fully observed windows. Coarse navigation commands follow the benchmark's future-ground-truth-displacement convention.
\begin{table}[!htbp]
\centering\footnotesize
\setlength{\tabcolsep}{3pt}
\caption{\textbf{Information-flow endpoint metrics.} L2 and collision C use all 6,019 validation windows.}
\label{tab:flow-details}
\begin{tabular*}{\columnwidth}{@{\extracolsep{\fill}}llrrr@{}}\toprule
Variant & Metric & 1\,s & 2\,s & 3\,s \\
\midrule
Single-frame context & L2 (m) & 0.2076 & 0.5252 & 1.0594 \\
 & C (\%) & 0.2492 & 0.4984 & 0.9802 \\
\ours{}: clean history & L2 (m) & 0.1869 & 0.4713 & 0.9388 \\
 & C (\%) & 0.4486 & 0.5150 & 0.7642 \\
Imagine then execute & L2 (m) & 0.3933 & 0.9968 & 1.8249 \\
 & C (\%) & 0.4486 & 0.6812 & 1.1962 \\
\bottomrule\end{tabular*}
\end{table}

\begin{table}[!htbp]
\centering\footnotesize
\setlength{\tabcolsep}{3pt}
\caption{\textbf{Complete-future sensitivity on 5,119 windows.} L2 and C average endpoints; C$_3$ is collision at 3\,s.}
\label{tab:flow-complete}
\begin{tabular*}{\columnwidth}{@{\extracolsep{\fill}}lrrr@{}}\toprule
Variant & L2 (m) & C (\%) & C$_3$ (\%) \\
\midrule
Single-frame context & 0.6829 & 0.6577 & 1.1526 \\
\ours{}: clean history & 0.6091 & 0.6381 & 0.8986 \\
Imagine then execute & 1.2317 & 0.8726 & 1.4065 \\
\bottomrule\end{tabular*}
\end{table}

\subsection{Runtime and delay evaluations}
\label{sec:supp-latency}
\subsubsection{Decision latency}

With world updates on a second H20, foreground execution sets decision latency: over 200 calls, decisions take \textbf{102.3}$\,\pm\,$3.6\,ms on NAVSIM and \textbf{109.9}$\,\pm\,$2.9\,ms on Bench2Drive without scoring, and \textbf{198.4}$\,\pm\,$3.2\,ms with it.
Relative to a retrained single-GPU synchronous baseline without a carrier (927.6\,ms, 93.9 PDMS), asynchronous decoupling runs $4.7\times$ faster at matched accuracy (94.2 and 94.0 PDMS at 1.0\,s and 1.5\,s ages).
A second GPU alone cannot remove the 760.6\,ms world-pass wait in synchronous designs, while reusing each update across two NAVSIM steps cuts two-GPU compute from 927.6 to 578.7\,ms ($1.6\times$; \cref{sec:supp-timing,sec:supp-sync}).

\subsubsection{Foreground and streaming world timing}
\label{sec:supp-timing}
\paragraph{Foreground timing.}
We profile 200 CUDA-synchronized decisions following 20 warmup calls on a dedicated NVIDIA H20 (GPU 0), while the world model publishes $\mathcal C_s$ concurrently from a second H20 (GPU 1). Each timed interval spans current-image host-to-device transfer, freshness carrier encoding, Action DiT forward pass and ODE integration, $\SEtwo$ rollout, and trajectory return to CPU.
At $\NODE=1$, seven-view NAVSIM with \TAEW{} averages \textbf{102.3\,ms} without scoring ($K=1$), and generating and ranking all $K=53$ candidates with the four-layer 5.2M scorer averages \textbf{198.4\,ms}; six-view Bench2Drive with \DINOvthree{} ($K=1$, fixed seed, no scorer) averages \textbf{109.9\,ms}.
Foreground compute totals $\Fdec=\Fcar+\NODE K\Fa+\Fsc$: relative to a full-history world pass $\Fw$, lightweight carriers cost 0.13--0.34\% FLOPs, each single-candidate action evaluation costs 0.035\% ($\approx 1.9\%$ for 53 one-step candidates), and candidate scoring adds $\approx 0.04\%$.

\paragraph{Streaming world timing.}
Both H20 GPUs execute with \texttt{torch.compile} and CUDA graphs using NVLink P2P publication: GPU 1 runs the causal \WanVAE{} and 28-layer 2B Video DiT, while GPU 0 runs the carrier and 28-layer 0.3B Action DiT. The world update latency $\tauw$ covers causal VAE state updates, newest-slot encoding, cache maintenance, and inter-GPU publication.
Across eight post-warmup streaming updates, a NAVSIM update takes \textbf{760.6}$\,\pm\,$2.0\,ms (mean $\pm$ sample std; maximum 764.0\,ms), and Bench2Drive's six $288\times512$ views take \textbf{941.5}$\,\pm\,$1.5\,ms (maximum 945.0\,ms, including 265.7\,ms for the VAE and 606.9\,ms for the Video DiT, leaving a 55\,ms margin before the next update is due).
With $\Tupd=1.0$\,s and $\Tpred=2.0$\,s, both benchmarks satisfy the background non-queuing and coverage conditions of \cref{eq:m-age}: $\tauw\le\Tupd$ and $\Tupd+\tauw\le\Tpred$, where $\Tupd+\tauw$ equals 1.76\,s on NAVSIM and 1.94\,s on Bench2Drive. Quantizing $\tauw$ to whole decision steps yields $\taud=1.0$\,s on both benchmarks (the chunk-step age in \cref{tab:timing}), from which \cref{eq:context-age} gives subsequent freshness-step ages.

\subsubsection{Synchronous and asynchronous execution}
\label{sec:supp-sync}
The single-GPU synchronous baseline retains the exact Video DiT, Action DiT, scorer, 10M-clip initialization, and training hyperparameters, but removes the freshness carrier and is retrained at $a_t=0$: because the Action DiT waits for each world update on one H20, its world context already contains the current images, and its 927.6\,ms latency spans the world pass, 53-candidate generation, and scoring in series without inter-GPU transfer. In \cref{tab:supp-sync}, both asynchronous rows evaluate a single deployed model (trained evenly across the 1.0\,s and 1.5\,s scheduled ages, as on Bench2Drive) by independently encoding each scene at the respective age while profiling foreground latency alongside concurrent world updates on the second H20.
Averaged across the two scheduled ages, \ours{} matches synchronous accuracy (94.1 mean PDMS vs.\ 93.9 synchronous; 90.7 and 90.9 unscored at the chunk and freshness steps) while cutting decision latency by $4.7\times$, and amortizing one world update across two decisions reduces total two-GPU compute per decision from 927.6\,ms to 578.7\,ms.
\begin{table}[!htbp]
  \caption{\textbf{Synchronous and asynchronous NAVSIM execution.} PDMS with the scorer and H20 decision latency. Both asynchronous rows evaluate the same model.}
  \label{tab:supp-sync}
  \centering
  \footnotesize
  \setlength{\tabcolsep}{0pt}
  \begin{tabular*}{\columnwidth}{@{\extracolsep{\fill}}lccc@{}}
    \toprule
    Execution & GPUs & \shortstack{Latency (ms)\\$\downarrow$} & PDMS $\uparrow$ \\
    \midrule
    Synchronous, $a_t{=}0$ & 1 & 927.6 & 93.9 \\
    Asynchronous, $a_t{=}1.0$\,s (chunk) & 2 & 198.4 & \textbf{94.2} \\
    Asynchronous, $a_t{=}1.5$\,s (freshness) & 2 & 198.4 & 94.0 \\
    \bottomrule
  \end{tabular*}
\end{table}

\begin{table}[!htbp]
\centering\footnotesize
\setlength{\tabcolsep}{6pt}
\caption{\textbf{Bench2Drive under injected delay.} Commands take effect after each delay as the scene advances; 0\,ms is the standard protocol. Mean $\pm$ sample std over three seeds.}
\label{tab:latency-injection}
\begin{tabular*}{\columnwidth}{@{\extracolsep{\fill}}rrr@{}}\toprule
Delay (ms) & DS $\uparrow$ & SR (\%) $\uparrow$ \\
\midrule
0 & $80.30\pm0.32$ & $56.36\pm0.79$ \\
100 & $75.62\pm0.58$ & $52.12\pm1.14$ \\
500 & $50.24\pm1.66$ & $16.67\pm1.84$ \\
1000 & $11.32\pm0.50$ & $0.91\pm0.45$ \\
\bottomrule\end{tabular*}

\end{table}

\subsubsection{Injected delay in closed loop}
To emulate on-vehicle deployment where actuators receive commands only after inference finishes, we inject control delays on Bench2Drive (\cref{tab:latency-injection}): each decision reads observations and cached world context from its trigger time while CARLA advances under prior commands.
At 100\,ms, matching our 109.9\,ms foreground latency, drop is modest (75.62 DS, 52.12\% SR).
At 500\,ms DS falls sharply to 50.24, and at 1000\,ms, near the 941.5\,ms world update of a synchronous pass, driving nearly collapses (11.32 DS, 0.91\% SR), confirming closed-loop driving requires ${\sim}100$\,ms latency and validating our fast foreground path (\cref{sec:supp-delay}).

\subsubsection{Delay-injected closed-loop evaluation}
\label{sec:supp-delay}
After initial warm-up, each decision at simulator timestamp $t_0$ consumes the camera images (including freshness inputs), ego state, and world cache available at $t_0$, while its resulting control command is applied at $t_0+d$ under injected delay $d\in\{100,500,1000\}$\,ms. Throughout the delay window, CARLA advances continuously while holding the previously applied throttle, brake, and steering commands, so subsequent observations reflect the vehicle's resulting motion. The next decision triggers once the delayed control takes effect, making $d$ the effective decision period: at $d=1000$\,ms, the planner decides at 1\,Hz with a 1\,s actuation lag, approximating the measured 941.5\,ms world-update latency. The $0$\,ms row corresponds to the standard protocol. All rows hold model weights fixed across three evaluation seeds of 220 routes each (\cref{tab:b2d-seeds}); \cref{tab:latency-injection} reports seed means and sample standard deviations (denominator $3-1$), with SR statistics computed from unrounded route-completion fractions. Out of $660$ route runs per delay, completed routes total 372, 344, 110, and 6 at delays of 0, 100, 500, and 1000\,ms, respectively. These probes quantify closed-loop sensitivity to control latency.
\begin{table}[!htbp]
\centering\footnotesize
\setlength{\tabcolsep}{4pt}
\caption{\textbf{Bench2Drive results for each evaluation seed.} Rows vary injected delay and seed. Columns report DS, SR, and successful routes out of 220.}
\label{tab:b2d-seeds}
\begin{tabular*}{\columnwidth}{@{\extracolsep{\fill}}rrrrr@{}}\toprule
Delay (ms) & Seed & DS & SR (\%) & Successes \\
\midrule
0 & 1 & 80.58 & 56.82 & 125 \\
0 & 2 & 79.95 & 55.45 & 122 \\
0 & 3 & 80.38 & 56.82 & 125 \\
\midrule
100 & 1 & 74.98 & 50.91 & 112 \\
100 & 2 & 76.13 & 53.18 & 117 \\
100 & 3 & 75.74 & 52.27 & 115 \\
\midrule
500 & 1 & 51.78 & 18.64 & 41 \\
500 & 2 & 50.46 & 16.36 & 36 \\
500 & 3 & 48.49 & 15.00 & 33 \\
\midrule
1000 & 1 & 11.23 & 0.91 & 2 \\
1000 & 2 & 10.87 & 0.45 & 1 \\
1000 & 3 & 11.86 & 1.36 & 3 \\
\bottomrule\end{tabular*}
\end{table}

\subsubsection{Comparator timing sources}
\label{sec:supp-comparator-timing}
On H20 GPUs, our scored NAVSIM decision (198.4\,ms) runs $1.7\times$ and $3.6\times$ faster than the paper-reported latencies of SV-WAM~\cite{svwam} (341.6\,ms) and DriveLaW~\cite{drivelaw} (710.0\,ms) at higher PDMS (\cref{tab:navsim}), whereas single-view PerceptDrive~\cite{perceptdrive} reports 68.0\,ms on an MI308X GPU at 90.4 PDMS.
\Cref{tab:supp-latency-evidence} documents the timing sources for \cref{tab:navsim,tab:b2d}: unmarked entries use each paper's reported hardware and operating point (converting FPS where needed, and summing Raw2Drive's module times as $600+2+2=604.0$\,ms~\cite{raw2drive}); daggered entries ($\dagger$) are end-to-end measurements from official code on one H20 (20 warmups, 200 calls). Our two-GPU measurements pair foreground planning on GPU 0 with concurrent world updates on GPU 1.

\begin{table}[!htbp]
  \caption{\textbf{Comparator timing sources.} A dagger denotes our official-code H20 rerun; n/s denotes an unspecified device.}
  \label{tab:supp-latency-evidence}
  \centering
  \footnotesize
  \setlength{\tabcolsep}{0pt}
  \begin{tabular*}{\columnwidth}{@{\extracolsep{\fill}}lccc@{}}
    \toprule
    Method & ms & GPU & Operating point \\
    \midrule
    \multicolumn{4}{@{}l}{\emph{NAVSIM comparators}} \\
    DriveLaW~\cite{drivelaw} & 710.0 & H20 & 5-step planning \\
    SV-WAM~\cite{svwam} & 341.6 & H20 & 2-step online total \\
    PerceptDrive~\cite{perceptdrive} & 68.0 & MI308X & 25-step end-to-end \\
    4D-WAM~\cite{fourdwam} & 790.0 & 4090 & 10-step default plan \\
    UNIVERSE~\cite{universe} & 376.0 & n/s & trajectory inference \\
    SimWAM~\cite{simwam} & 518.0 & A100 & 10-step, imitation model \\
    GoalFlow~\cite{goalflow} & $532.1^{\dagger}$ & H20 & 5-step, full decision \\
    ReCogDrive~\cite{recogdrive} & 75.0 & n/s & VLM + 5-step plan \\
    Hydra-MDP++~\cite{hydramdppp} & 271.0 & V100 & V2-99 backbone \\
    DriveSuprim~\cite{drivesuprim} & 80.0 & n/s & ViT-L, 12.5\,FPS \\
    \midrule
    \multicolumn{4}{@{}l}{\emph{Bench2Drive comparators}} \\
    MindDriver~\cite{minddriver} & 1000.0 & 4090 & 1\,Hz system cycle \\
    Raw2Drive~\cite{raw2drive} & 604.0 & n/s & $600{+}2{+}2$\,ms modules \\
    UniDrive-WM~\cite{unidrivewm} & 2500.0 & A100 & 0.4\,FPS joint cycle \\
    \bottomrule
  \end{tabular*}
\end{table}

\section{Discussion}
\label{sec:discussion}

\subsection{Autoregressive imagination and generative world knowledge}
\label{sec:supp-generative}
\begin{figure*}[t]
  \centering
  \includegraphics[width=\textwidth]{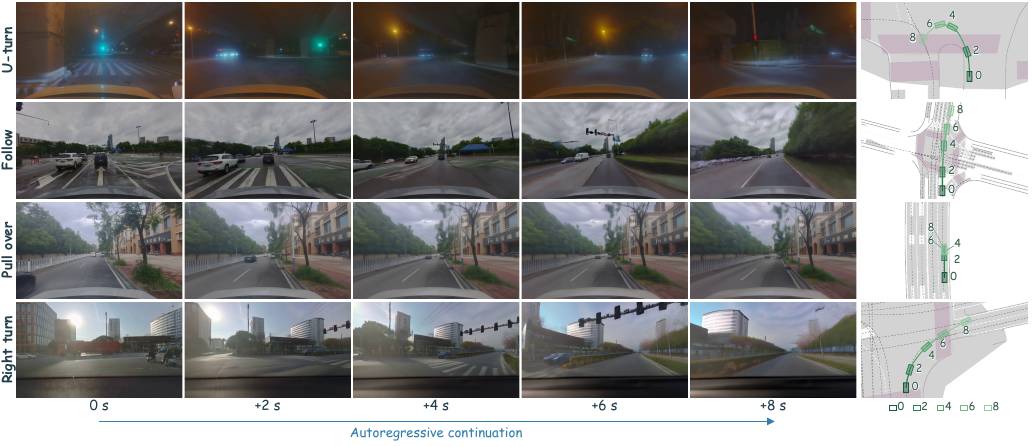}
  \input{figures/captions/fig_ar_caption.tex}
\end{figure*}

In \cref{fig:ar}, starting from 29 observed frames at $t\le0$, our Video DiT pretrained jointly on 10M clips autoregressively generates roughly 8\,s of coherent front-view video without future ego-pose, action, or navigation-command conditioning, spanning a nighttime U-turn under an overpass, urban intersection following, pulling over, and an unprotected right turn. Across these qualitative rollouts, planning over two successive imagined 4\,s windows produces smooth, road-compliant 8\,s ego trajectories beyond NAVSIM's 4\,s horizon.

Quantitatively on NAVSIM without scoring ($K=1$), keeping the public Cosmos-Predict2.5 Video DiT frozen yields 83.6 PDMS, whereas joint training on \texttt{navtrain} (updating $\Wtheta$ with both the video flow loss and action gradients through $\mathcal C_s$) reaches 88.5 PDMS (+4.9 points under the identical target-data setting), and scaling joint pretraining to 10M clips reaches 90.7 PDMS.
Together, these results link larger-scale generative learning to predictive world representations that the fast planner can reuse between background updates, alongside current visual evidence.

\subsection{Relation to asynchronous world-action models}
\label{sec:supp-async-comparison}
\begin{table*}[t]
  \caption{\textbf{Asynchronous world-action designs.} Entries follow each paper's description. Views counts camera inputs; the last column gives the path by which current observations reach action steps between slow updates.}
  \label{tab:async-comparison}
  \centering
  \footnotesize
  \begin{tabular*}{\textwidth}{@{\extracolsep{\fill}}lllll@{}}
    \toprule
    Method & Domain (views) & Reused slow context & Cache between updates & Current observations \\
    \midrule
    AHA-WAM~\cite{ahawam} & Manip. (1--3) & Planner K/V & Adapted per chunk & Routing queries from current view \\
    GlanceWAM~\cite{glancewam} & Manip. (2--3) & Generated lookahead latent & Held & Video DiT on current frame \\
    DualWAM~\cite{dualwam} & Manip. (global, wrist) & Partially denoised plan & Refined per window & Wrist-view WAM \\
    ZimaBlue~\cite{zimablue} & Manip. (3) & K/V from video-action rollout & Reused unchanged & Observation latents in fast DiT \\
    \ours{} & Driving (6--7) & K/V of clean history & Read-only & Lightweight encoder, own K/V bank \\
    \bottomrule
  \end{tabular*}
\end{table*}

\Cref{tab:async-comparison} summarizes context reuse and visual feedback in asynchronous world-action systems. AHA-WAM adapts cached planner K/V at each action chunk through routing queries. GlanceWAM holds a generated lookahead latent and runs the Video DiT on current frames at every control step. DualWAM refines a partially denoised global plan with a wrist-view WAM, and ZimaBlue's fast DiT reads an unmodified layer-wise K/V cache between slow updates. In \ours{}, the slow branch streams clean history into reusable K/V; the fast branch reads this cache alongside live features from a frozen lightweight encoder and a dedicated gated K/V bank. Across six to seven surround views, the first-read delay $\taud$ follows measured world update latency $\tauw$, and the trained predictive span covers the update interval plus that latency (\cref{eq:context-age,eq:m-age}).

\section{Conclusion}
\label{sec:conclusion}

We presented \textbf{\ours{}}, a deployment-oriented asynchronous world-action system for autonomous driving that unites a background 2B Video DiT for \emph{slow world} context, a lightweight carrier for \emph{instant freshness}, and a foreground 0.3B Action DiT for \emph{timely action}.
Training the predictive span to cover the cache reuse window while injecting live surround views at every step plans $4.7\times$ faster than synchronous execution at matched accuracy, leading compared world-action models across NAVSIM, Bench2Drive (including under command delay), and Waymo E2E.
\textbf{Limitations.} Concurrent execution currently requires two GPUs and a brief stationary cold start, motivating single-GPU distillation and on-vehicle validation.

\section*{LLM Usage Statement}
Claude Code (Anthropic) and OpenAI Codex were used for text editing and grammar polishing, to compile the comparison tables from the cited papers and cross-check them against the source papers, and to write the scripts that plot our results from our data.
The authors are responsible for the method, model and experiment code, experimental verification, and references, and take full responsibility for the content of this paper.

\FloatBarrier
\end{document}